# Automatic item generation for personality situational judgment tests with large language models

Chang-Jin Li [a,1], Jiyuan Zhang [a,1], Yun Tang [b,c], Jian Li [a,d,e,*]

[a] Faculty of Psychology, Beijing Normal University, Beijing, 100875, China
[b] School of Psychology, Central China Normal University, Wuhan, Hubei, 430079, China
[c] Key Laboratory of Adolescent Cyberpsychology and Behavior, Ministry of Education, Wuhan, Hubei, 430079, China
[d] Beijing Key Laboratory of Applied Experimental Psychology, Beijing, 100875, China
[e] Research Center for Capacity Building in Educational Assessment and Evaluation (Beijing Higher Education Innovation Center for Philosophy and Social Sciences), China

ARTICLE INFO

Dataset link: Replication Materials for: Automatic Item Generation for Personality Situational Judgment Tests with Large Language Models (Original data)

ABSTRACT

Personality assessment through situational judgment tests (SJTs) offers unique advantages over traditional Likert-type self-report scales, yet their development remains labor-intensive, time-consuming, and heavily dependent on subject matter experts. Recent advances in large language models (LLMs) have shown promise for automatic item generation (AIG). Building on these developments, the present study focuses on developing and evaluating a structured and generalizable framework for automatically generating personality SJTs, using GPT-4 and ChatGPT-5 as empirical examples. Three studies were conducted. Study 1 systematically compared the effects of prompt design and temperature settings on the content validity of LLM-generated items to develop an effective and stable LLM-based AIG approach for personality SJT. Results showed that optimized prompts and a temperature of 1.0 achieved the best balance of creativity and accuracy on GPT-4. Study 2 examined the cross-model generalizability and reproducibility of this automated SJT generation approach through multiple rounds. The results showed that the approach consistently produced reproducible and high-quality items on ChatGPT-5. Study 3 evaluated the psychometric properties of LLM-generated SJTs covering five facets of the Big Five personality traits. Results demonstrated satisfactory reliability and validity across most facets, though limitations were observed in the convergent validity of the compliance facet and certain aspects of criterion-related validity. These findings provide robust evidence that the proposed LLM-based AIG approach can produce culturally appropriate and psychometrically sound SJTs with efficiency comparable to or exceeding traditional methods. Beyond reducing expert dependency and development time and cost, this approach demonstrates how LLM-based methodologies can meaningfully expand AIG of SJTs and streamline test development processes in resource-limited contexts.

## 1. Introduction

Situational Judgment Tests (SJTs) are psychological assessment methods that present respondents with realistic scenarios and multiple response options to evaluate how they would likely behave in a given situation (Slaughter et al., 2014). Unlike traditional Likert-type self--report scales, SJTs simulate real-world contexts and capture the interaction between individual traits and situational demands, making them a valuable tool for personality assessment (Olaru et al., 2019). Prior research has further demonstrated that SJTs are less vulnerable to social desirability and deliberate falsification (Hooper et al., 2006; Kasten et al., 2020), more capable of reducing cognitive load and eliciting engagement (Lievens et al., 2008), and has shown reduced adverse impact on minority groups (Lievens et al., 2008) compared to traditional Likert-type scales. To illustrate the format of personality SJT, we include below an example item developed by Mussel et al. (2018) in German. This particular item is designed to assess self-consciousness, a facet of the Big Five factor neuroticism. The item was translated into English using DeepL, an AI-powered machine translation tool, and is presented only for illustrative purposes; no psychometric revision or validation of the English version was undertaken. Importantly, we obtained permission from Mussel and his colleagues to use and translate these items for the purposes of the present study.

* Corresponding author. Faculty of Psychology, Beijing Normal University, Beijing, 100875, China.
*E-mail address:* jianli@bnu.edu.cn (J. Li).

[1] Chang-Jin Li and Jiyuan Zhang contributed equally to the current research and should be considered joint first authors.

Example SJT item:

You are giving a presentation in front of your department colleagues. While you are speaking, you notice two female colleagues suddenly start laughing and whispering to each other. What would you do?

A I would wonder if I said something funny and look down to check if my clothes are impeccable.
B I would lose my train of thought and have to check my notes.
C I would continue with my presentation.
D I would pause briefly and ask the two female colleagues if anything was unclear or hard to understand in my presentation.

Scoring: Options A and B represent behavioral responses typically associated with individuals high in self-consciousness, therefore scored as 1 point; options C and D reflect behavioral tendencies associated with lower levels of self-consciousness, therefore scored as 0 point.

Despite these advantages, developing SJTs remains a labor-intensive and time-consuming process. Constructing realistic and psychometrically sound scenarios requires multiple stages, including the identification of critical incidents, drafting of situational vignettes, and generation of plausible response options, all of which demand substantial involvement from subject matter experts (SMEs; Corstjens et al., 2017). Moreover, these steps rely heavily on SMEs' personal experience and perspectives. Such reliance introduces subjectivity, inconsistency, and bias and error into the item development process (Rush et al., 2016).

The emergence of large language models (LLMs) offers new opportunities to address these limitations. LLMs can generate highly contextual, semantically rich scenarios, reducing the dependence on SMEs and accelerating the development process (Krumm et al., 2024). Recent studies have shown that LLM-generated scales can achieve comparable levels of reliability and validity to expert-developed ones, while significantly lowering time and resource costs (e.g., Götz et al., 2024; Hernandez & Nie, 2023; Hommel et al., 2022; Krumm et al., 2024; P. Lee et al., 2023). Nevertheless, most prior studies have focused on applying LLMs to the generation of traditional Likert-type personality scales. To date, only a few peer-reviewed studies have examined the validity of SJTs generated by LLMs. Krumm et al. (2024) provided initial evidence using GPT-3.5, yet their investigation was limited in scope: it generated items for a single facet—the gregariousness facet of the extraversion domain—and tested only their internal consistency and construct validity. More recently, Zhang et al. (2026) expanded this line of research by generating SJTs for the comprehensive HEXACO model using multiple LLMs (e.g., GPT-4, Claude 3.5) and exploring chain-of-thought prompting. However, neither study systematically examined the content validity of generated items across different parameter settings and multiple prompt strategies or evaluated the temporal stability of the generation process over time. Consequently, questions regarding the optimal configuration and reproducibility of LLM-based SJT generation remain largely unanswered.

In the current study, we sought to develop and evaluate a structured and generalizable methodology for automated personality SJT generation using LLMs, with the goal of producing items that demonstrate satisfactory reliability, validity, and reproducibility, thereby addressing key unanswered questions of the LLM-based SJT development process.

## 2. Literature review

### 2.1. Automatic item generation

Automatic item generation (AIG) is a technique that has gained increasing attention for its ability to streamline the test development process. Initially proposed in the 1970s, AIG has recently gained wider adoption, primarily due to advancements in computational power and algorithm design (Li & Zhang, 2008). AIG uses specially designed algorithms to generate test items based on specific theoretical frameworks and psychometric principles, ensuring each item is customized to meet defined parameters (Embretson, 2005; Embretson & Yang, 2007). Unlike manually-generated test items, which are often time-consuming and resource-intensive, AIG can quickly produce large volumes of items tailored to various test-taker abilities and needs.

One of AIG's key advantages is its capacity to enhance the efficiency and quality of test development. By specifying psychometric properties such as difficulty levels, AIG reduces the need for traditional trial-and-error processes, where poorly performing items are often discarded after pilot testing. Additionally, the automatic process ensures structural validity at the item level, allowing for more precise control over cognitive complexity (Embretson, 2005). This enhances the overall accuracy and reliability of the tests.

However, despite its strengths, traditional AIG approaches like item modeling and cognitive design system approaches have limitations when generating non-cognitive assessments, such as personality tests. The item modeling approaches rely on templates to create new items by replacing irrelevant elements (e.g., names, numbers) while maintaining the item's core structure (Bejar et al., 2002; Gierl & Lai, 2012). The cognitive design system approaches construct items based on mental models, mainly focusing on identifying critical features of task-solving processes (Bejar et al., 2002). While these approaches excel in measuring cognitive functions like mathematics (Gierl & Lai, 2015) and spatial reasoning (Arendasy, 2005; Arendasy et al., 2010), where items can be modeled using clear, well-defined rules, they are less suited to generating items for personality assessments, especially situational judgment tests (SJTs) that require more nuanced, context-rich scenarios that capture complex human behaviors and personality traits (Hommel et al., 2022).

Several approaches to AIG have been developed to address these challenges, such as semantic analysis and deep learning approaches. Semantic analysis approaches analyze the syntax and vocabulary of existing items to generate new ones. Nevertheless, these methods rely on template-like frameworks, limiting their ability to create the rich, context-specific items necessary for SJTs (Huang & He, 2016). Deep learning approaches, such as recurrent neural networks (RNNs) or transformer-based models like GPT, promise to overcome these limitations. These models can generate items based on large text corpora, producing more varied and nuanced items (von Davier, 2018). While deep learning approaches to AIG have brought significant advancements to cognitive testing (e.g., Attali et al., 2022; Doughty et al., 2024), its application in non-cognitive domains like personality assessment remains challenging. These advanced methods require further refinement, particularly in distinguishing between specific constructs and maintaining the psychometric quality of generated items (Hernandez & Nie, 2023; Hommel et al., 2022).

### 2.2. Research on generating personality tests using large language models

Trained on vast text corpora, LLMs offer significant advantages over traditional AIG methods. Unlike template-based or cognitively modeled AIG approaches, LLMs excel in generating semantically rich and contextually diverse content (Demszky et al., 2023), making them ideal for generating non-cognitive assessments like personality tests. Their ability to capture nuanced language and behavior is especially valuable for creating SJTs that mirror complex interpersonal dynamics and real-world scenarios.

The application of LLMs in personality test generation has shown promising potential. Hommel et al. (2022) demonstrated that GPT-2 could generate items comparable to manually crafted ones, though internal consistency remained a challenge, highlighting the need for further fine-tuning. Hernandez and Nie (2023) improved GPT-2's performance through iterative fine-tuning, achieving high Cronbach's alpha coefficients and maintaining traditional factor structures, though issues with grammatical errors persisted. By contrast, Götz et al. (2024) found that most items generated by a cloned GPT-2 model failed to meet psychometric criteria, emphasizing the limitations of insufficient fine-tuning. Later, research on GPT-3 revealed its superior ability to capture personality nuances, with prompt-based item generation producing favorable psychometric properties (P. Lee et al., 2023). Recently, Krumm et al. (2024) examined the psychometric properties of an SJT developed by GPT-3.5 in comparison to a human-created SJT, both measuring gregariousness as a personality facet. Their study revealed that while the initial 11-item ChatGPT-created SJT exhibited poor model fit, a refined 8-item version achieved acceptable fit. The ChatGPT-created SJT demonstrated comparable internal consistency ($\omega = 0.82$) and convergent validity ($r = 0.64$ with self-reported gregariousness) to the human-developed SJT ($\omega = 0.68$, $r = 0.69$). Expanding this line of research, Zhang et al. (2026) utilized multiple LLMs to generate SJTs for the HEXACO personality model. Their results indicated that, overall, LLM-generated items demonstrated internal consistency and convergent validity comparable to human-developed items, further validating the feasibility of LLMs in broader personality frameworks.

Despite these advancements, several challenges remain in generating high-quality personality test items using LLMs. One significant issue is the balance between scenario diversity and psychometric quality (Krumm et al., 2024). While LLMs can generate diverse and context-rich items, they sometimes produce text that lacks the coherence or precision necessary for psychometric assessments. The variability in LLM output is often linked to the settings of temperature parameters, which control the randomness of the generated text (Demszky et al., 2023). A higher temperature increases the likelihood of the model selecting lower-probability words, enhancing the output's creativity and diversity (OpenAI et al., 2024). However, if the temperature is set too high (e.g., higher than 1.0), the output may become excessively disorganized and difficult to interpret, ultimately reducing text quality. Conversely, lower temperatures produce more conservative and repetitive outputs, limiting the diversity needed for SJT. When the temperature is set to 0, the LLM completely loses randomness, deterministically selecting the highest-probability token. Therefore, finding the optimal temperature parameter is critical for developing personality scales. According to the API reference (n.d.) released by OpenAI, the default temperature parameter value for chat completion is 1.0, within a range of 0 to 2. Prior research indicates that a temperature setting in the range of 0.7 to 1.0 is recommended when conducting AIG with LLMs (Götz et al., 2024). Striking the balance ensures that the LLM generates items that are both sufficiently varied and psychometrically sound, avoiding excessive randomness while preventing overly generic formulations.

Another critical challenge is ensuring that the items generated by LLMs accurately reflect the intended psychological constructs. Götz et al. (2024) found that although LLMs could generate vast quantities of items, many of these items lacked precise alignment with the underlying constructs, leading to poor psychometric performance. Zhang et al. (2026) observed that items targeting Honesty-Humility, Emotionality, Agreeableness, and Openness to Experience exhibited weaker psychometric quality compared to traits like Extraversion and Conscientiousness, suggesting that LLMs still face difficulties in operationalizing complex, socially grounded traits.

To address this challenge, recent research has focused on prompt engineering—carefully designing input prompts to guide LLMs in generating high-quality outputs (U. Lee et al., 2024; Oeljeklaus et al., 2025; Wang et al., 2024; Zhang et al., 2026). Prior studies have noted that insufficiently structured prompts often undermine the conceptual clarity and formatting consistency of generated Likert-type personality items (P. Lee et al., 2023). The syntax (e.g., length, whitespace, order of examples) and semantics (e.g., wording, instructions, example selection) of prompts can significantly impact the quality of LLM outputs (Marvin et al., 2023).

## 3. The present study

Recent progress in AIG has highlighted both the opportunities and limitations of using LLMs to generate SJT items. Although previous studies have demonstrated the potential to generate diverse and high-quality content with LLMs, persistent challenges remain, particularly in conceptual relevance and linguistic clarity. For instance, while ChatGPT-created SJT has been found to mirror human-created SJT in terms of internal consistency reliability and convergent validity, the presence of multidimensionality of scenarios and misclassification of response options remains a significant challenge (Krumm et al., 2024). Furthermore, prior studies have not thoroughly examined whether LLM-generated SJT items maintain adequate content validity across variations in parameter settings, prompt strategies, generation time points, or different models. Against this backdrop, the present study undertakes an exploratory investigation aimed at developing and evaluating a more structured and generalizable approach to LLM-based SJT generation. The focus of the study lies in examining how methodological components shape the quality of LLM-generated SJT items.

To systematically address these objectives, the present research followed a three-study progressive design, as illustrated in Fig. 1. Study 1 aimed to develop an effective and stable LLM-based AIG approach for SJT. It focused on technical optimization by identifying the most effective prompt strategies and temperature settings to maximize the content validity of LLM-generated items. Study 2 utilized these optimized parameters to evaluate the cross-model generalizability and reproducibility of the LLM-based AIG approach across broader personality domains and multiple generation rounds. Finally, Study 3 empirically validated the practical utility of the proposed methodology by subjecting the LLM-generated personality SJTs to a comprehensive psychometric evaluation, assessing their reliability and validity against traditional human-developed benchmarks.

## 4. Study 1: content validity and stability of LLM-generated personality situational judgment tests

Study 1 aimed to develop an effective and stable LLM-based AIG approach to SJT generation. To achieve this objective, the study was further designed to address three key research questions that clarify how different methodological choices influence the quality and stability of the generated items: (1) how temperature settings influence the content validity of LLM-generated SJT items, (2) how prompt strategies can be optimized to enhance the content validity of LLM-generated SJT items, and (3) do LLM-generated SJT items maintain stable content validity when generated at different times using identical prompts and parameters. In this study, GPT-4 was used to automatically generate SJT items focusing on the self-consciousness facet of the Big Five factor neuroticism. An expert panel assessed the content validity of the LLM-SJT items and the SJT items manually designed by Mussel et al. (2018).

This study involved three comparisons: (1) temperature variations, comparing the content validity of manually-generated SJT and LLM-generated SJT at various temperature settings; (2) prompt variations, comparing the content validity of manually-generated items and LLM-SJT items generated using different prompt strategies; (3) stability validation, comparing the content validity of LLM-SJT generated at different times using the same prompt strategies and temperature settings.

### 4.1. Participants

The expert panel consisted of seven doctoral students in psychology from Beijing Normal University, Central China Normal University, and Shaanxi Normal University, all of whom were familiar with the development of personality tests and situational judgment tests.

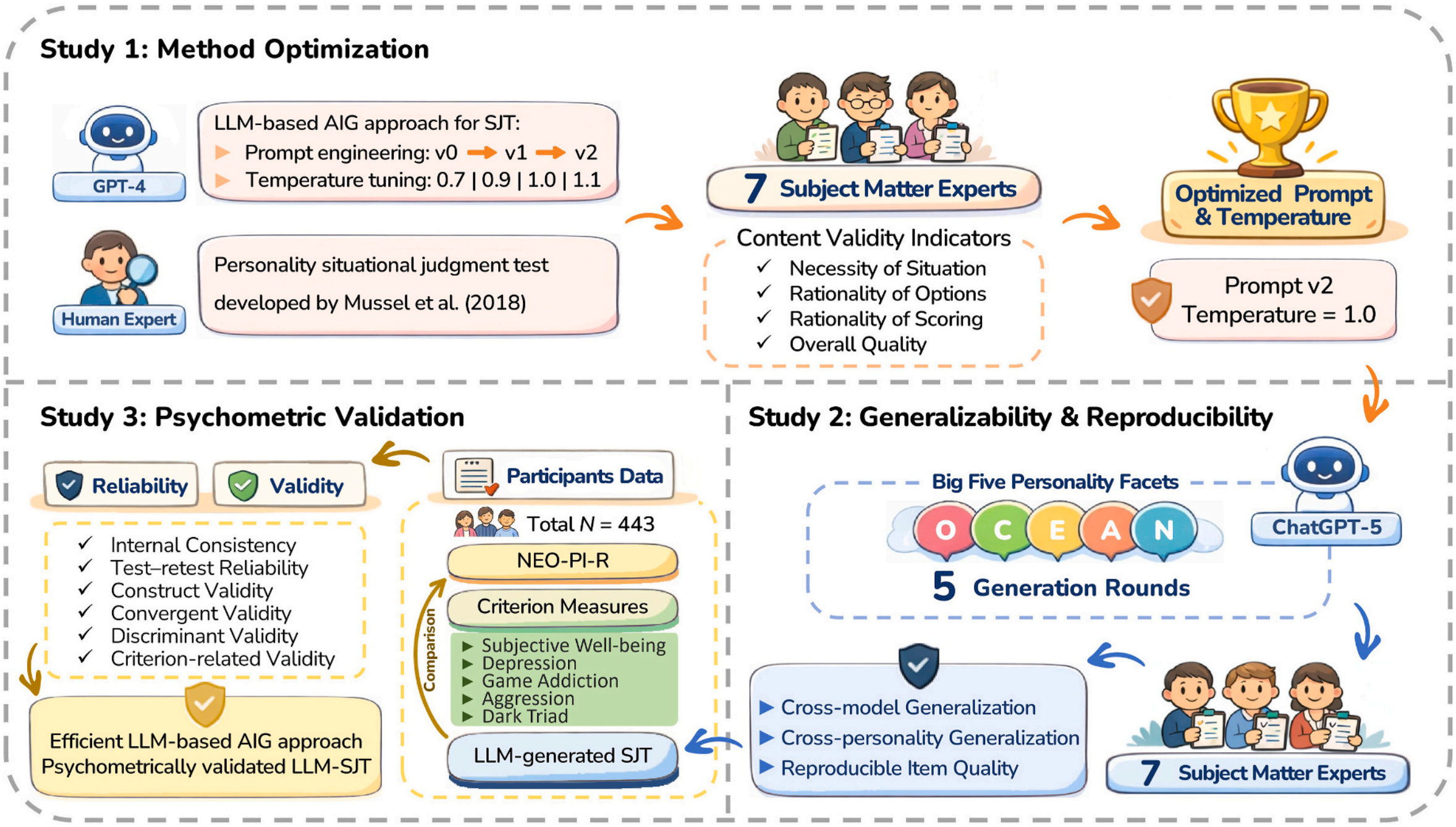

**Fig. 1.** Overview of the three-study pipeline.

## 4.2. Materials

### 4.2.1. Personality situational judgment test (Mussel et al., 2018)

The original German version of the test contains 110 items and uses the five facets of the Big Five personality instead of the five factors. The facets involved self-consciousness (from the Big Five factor neuroticism), gregariousness (from extraversion), openness to ideas (from openness to experience), compliance (from agreeableness), and self-discipline (from conscientiousness). Each facet contains 22 items, each offering four options—two representing a higher level of a specific trait (1 point) and two representing a lower level (0 points). In this study, 22 items on self-consciousness were used. One of the items was selected as an example of content validation scoring, with the remaining 21 items being labeled as Manually-generated Group (MG) and used for subsequent comparison with LLM-generated items.

Before being utilized in this study, the original German version of the test was first translated into Chinese. The German-Chinese adaptation followed a rigorous translation-back translation protocol to ensure conceptual and semantic equivalence. The original German items were first translated into Chinese using DeepL. The Chinese version was then back-translated to German through a new session of the same platform to avoid algorithmic memory bias. After that, a bilingual expert (native Chinese speaker) compared the back-translated German version with the original text to identify semantic discrepancies. Discrepancies were resolved through iterative revisions of the Chinese translation until achieving optimal conceptual alignment. Finally, a culturally adapted version was further refined by a psychometrician specializing in SJT development to ensure idiomatic fluency and contextual relevance for Chinese respondents while preserving the original meaning, achieving both linguistic accuracy and psychological equivalence across cultures.

### 4.2.2. Personality situational judgment test generated by GPT-4

SJT items for the personality facet of self-consciousness were generated by OpenAI's "gpt-4-1106-preview" model. Different temperature settings and prompts were tested to determine the optimal temperature settings and prompt strategies, with seven items generated for each condition. There was no item editing or selection conducted in the following procedures.

**Temperature Setting.** Previous studies have employed various temperature parameter settings, i.e. a temperature of 1.0 used by P. Lee et al. (2023), a temperature of 0.9 used by Götz et al. (2024), and temperatures of 0.7, 0.9, and 1.1 used by Hommel et al. (2022). In the current study, we tested temperature values of 0.7, 0.9, 1.0, and 1.1.

**Prompt Strategy.** Three versions of prompt strategies were employed with a temperature of 1.0, which is the default value for chat completion by OpenAI (API reference, n.d.). Prompt v0, adapted from the strategy used by P. Lee et al. (2023) for generating Likert-type personality scales, uses only repeated instructions, learning examples, and clear formatting, but lacks detailed explanations of construct meanings or precise output requirements. Prompt v1 enhances Prompt v0 by incorporating precise construct definitions and behavior descriptions, detailed context and output specifications, Chain-of-Thought reasoning, expert persona adoption, and emotional prompting, guiding LLMs to generate more diverse and context-rich items. Prompt v2 incorporated four key improvements—integrating the persona with task objectives, clarifying output requirements, removing redundant behavior descriptions, and optimizing examples—building on Prompt v1 to further enhance content validity. It should be noted that only Prompt v1 was used for temperature comparison experiments. The full text of Prompts v0, v1, and v2 can be found in online Supplementary Materials (https://osf.io/jbvq7/).

**Temporal Stability.** Two sets of test items were generated using the prompt v1 and a temperature of 1.0, with a 10-day interval, to examine the stability of content validity for LLM-SJTs generated at different time points using identical prompts and temperature settings.

## 4.3. Procedures

The LLM-generated items and manually-generated items were randomized and mixed. A panel of seven experts rated the content validity using four indicators. These include the necessity of the situation, the rationality of options, the rationality of scoring, and the overall evaluation of item quality (Christian et al., 2010; P. Lee et al., 2023). We explicitly assess the rationality of options and the rationality of scoring because, in SJT development, the options themselves are critical content. Each option represents a distinct behavioral response corresponding to different levels of the measured trait. However, prior research has shown that LLMs may fail to adhere to scoring rubrics—for instance, generating fewer high-trait options than requested or misclassifying options as high-trait despite incongruent content (Krumm et al., 2024). By including option evaluation, we ensure that both the scenarios and their associated responses meet psychometric standards. The instructions first clarified that the items include both manually-generated and LLM-generated items. Then, experts were provided a detailed definition of the self-consciousness facet of personality. They were subsequently asked to evaluate each item based on four indicators.

- The necessity of the situation: Experts are required to evaluate the extent to which an individual's behavior, as described in the scenario, reflects the level of the target personality traits. The rating scale has three levels: Necessary and useful, which indicates that the scenario is indispensable or crucial for reflecting the level of the personality traits (scored as 1); Useful but not necessary, which means the scenario is helpful but not essential (scored as 2); Neither necessary nor useful, meaning the scenario does not contribute to reflecting the level of the personality traits (scored as 3). Based on these ratings, the content validity ratio (CVR; Lawshe, 1975) was calculated using the formula CVR=$(n-N/2)/(N/2)$, where $N$ is the total number of experts and $n$ is the number of experts who consider the scenario necessary for reflecting the level of the target personality traits. The CVR ranges from −1 to +1, with positive values indicating that more than half of the experts agree that the context of the item is essential, and negative values indicating that fewer than half of the experts hold this view. Based on the criterion proposed by Baghestani et al. (2019), a minimum critical CVR of 0.71 is required for a 7-member expert panel. In addition to the item-level assessment using CVR, the overall content validity of the scale was evaluated using the Content Validity Index (CVI; Lawshe, 1975). The CVI, calculated as the mean of all item-level CVRs, provides a summary measure of the entire scale's validity. The critical value for the CVI was set at 0.70, following the guideline proposed by Tilden et al. (1990).
- The rationality of options: Experts were required to evaluate the rationality of each option. A reasonable option should be realistic and contextually relevant, meaning the behavior described could plausibly occur in the scenario. The rating scale is as follows: 0 for no options are reasonable, 1 for one option is reasonable, 2 for two options are reasonable, 3 for three options are reasonable, and 4 for four options are reasonable. The score ranges from 0 to 4. The average of the experts' ratings was used as the score of the rationality of options.
- The rationality of scoring: Experts are required to evaluate the rationality of scoring. An accurate scoring should award 1 point for options representing a high level of the trait and 0 points for options representing a low level of the trait. The rating scale is as follows: 0 for no options are scored accurately, 1 for one option is scored accurately, 2 for two options are scored accurately, 3 for three options are scored accurately, and 4 for four options are scored accurately. The score ranges from 0 to 4. The average of the experts' ratings was used as the score of the rationality of scoring.
- Overall quality: Experts are required to evaluate whether each item is suitable for directly measuring the target personality traits based on a comprehensive evaluation of its syntactic and linguistic correctness, contextual richness, psychological fidelity, and scoring method. No is scored as 0, and yes is scored as 1. The sum of the experts' ratings was used as the score of overall quality.

### 4.4. Statistical analysis

Despite implementing rigorous translation-back translation procedures and cultural adaptation protocols, certain items exhibited persistent sociocultural incongruence in the Chinese context. During the content validity evaluation, nearly half of the items were flagged as culturally inappropriate by three or more raters (overall quality ratings <5/7), with one particular item unanimously rejected (overall quality rating = 0). The problematic item reads: "You are invited to a birthday party and are sitting at a table with other guests, most of whom you do not know. A guest informs you that the host would be very pleased with a small impromptu birthday speech from you. How do you behave?" Raters particularly criticized options A ("I explain to the birthday person that I am reluctant to give a speech") and B ("I directly decline the suggestion and do not give a birthday speech, as I do not like speaking in front of so many people"). In Chinese culture, behaving in accordance with societal and interpersonal expectations during social interactions is essential for reinforcing *guanxi* ("关系", relationships; Qi, 2013), preserving *mianzi* ("面子", face; Wei et al., 2023), and upholding *he* ("和", harmony; Wei & Li, 2013). Declining such requests often incurs greater social pressure than reluctant compliance, contrasting with Western contexts where individual autonomy in social participation is culturally prioritized (Schwartz, 2014). To address potential cultural bias in comparing manually-generated and LLM-generated items, we implemented dual inclusion criteria for the translated manually-generated items: (1) the CVR for the necessity of the situation surpasses 0; (2) the overall evaluation of the quality achieves a minimum score of 5 out of 7. Finally, ten culturally adapted items were retained and labeled as MG-ca.

The first analysis compared the content validity of LLM-generated items using Prompt v1 at different temperature settings (labeled as Temp0.7, Temp0.9, Temp1.0, and Temp1.1) with items in MG and MG-ca, separately. A Kruskal-Wallis H test was conducted using IBM SPSS Statistics (Version 26) to assess whether there were statistically significant differences between the item groups in terms of the four indicators of content validity. If significant differences were found, Dunn's post hoc tests were performed with Bonferroni error correction.

The second analysis compared the content validity of LLM-generated items using different prompt strategies (labeled as Prompt v0, Prompt v1, and Prompt v2) at a temperature setting of 1.0 with items in MG and MG-ca, separately. The statistical methods and the indicators of content validity were the same as those used in the first analysis.

The third analysis compared the differences between items generated at two time points, which were 10 days apart, using Prompt v1 at a temperature setting of 1.0. Data for time point 1 were from the first analysis mentioned above, and data for time point 2 are from the second analysis. Using IBM SPSS Statistics (Version 26), a Mann-Whitney $U$ test was performed to determine whether there were statistically significant differences between the two item groups (labeled as Time1 and Time2) across the four indicators of content validity.

Additionally, we conducted an analysis of inter-rater reliability among expert ratings by calculating the intraclass correlation coefficient (ICC) using a two-way mixed-effects model for absolute agreement, ICC (2,k).

### 4.5. Results

#### 4.5.1. Inter-rater reliability of expert ratings

Inter-rater reliability analysis revealed moderate agreement in the necessity of the situation (0.63), the rationality of options (0.52), the rationality of scoring (0.68), and the overall evaluation of item quality (0.64), according to Koo and Li's (2016) suggested thresholds for moderate reliability (0.50 < ICC <0.75).

#### 4.5.2. Descriptive statistics of content validity across different temperature settings and prompt strategies

Table 1 presents the descriptive statistics for the four content validity indicators of manually-generated items, as well as across temperature settings and prompt strategies. Using the critical value of CVI ≥0.70 proposed by Tilden et al. (1990) as the criterion for acceptable content validity, clear differences emerged across conditions.

Across temperature settings, CVI values increased with higher temperature. While Temp0.7 and Temp0.9 remained below the 0.70 threshold, Temp1.0 and Temp1.1 both exceeded the criterion, producing higher CVI than the MG and MG-ca groups. Other content validity

**Table 1**
Means and standard deviations of content validity indicators across temperature settings and prompt strategies.

| Condition | Number of items | Necessity of the situation | Rationality of options | Rationality of scoring | Overall item quality |
|---|---|---|---|---|---|
| MG | 21 | 0.22 (0.60) | 3.59 (0.30) | 3.57 (0.31) | 4.00 (1.79) |
| MG-ca | 10 | 0.63 (0.27) | 3.76 (0.19) | 3.80 (0.17) | 5.40 (0.70) |
| Temp0.7 | 7 | 0.55 (0.32) | 3.92 (0.14) | 3.86 (0.12) | 5.71 (0.95) |
| Temp0.9 | 7 | 0.51 (0.49) | 3.82 (0.23) | 3.90 (0.14) | 5.86 (1.68) |
| Temp1.0 | 7 | 0.71 (0.44) | 3.96 (0.07) | 3.82 (0.23) | 5.71 (1.25) |
| Temp1.1 | 7 | 0.71 (0.16) | 3.84 (0.15) | 3.67 (0.36) | 5.29 (1.50) |
| Prompt v0 | 7 | 0.14 (0.47) | 3.71 (0.20) | 3.49 (0.62) | 4.86 (2.12) |
| Prompt v1 | 7 | 0.67 (0.35) | 3.90 (0.18) | 3.90 (0.14) | 6.57 (0.79) |
| Prompt v2 | 7 | 0.76 (0.26) | 3.90 (0.11) | 3.90 (0.21) | 6.00 (0.82) |

*Note.* Values in the last four columns are presented in the format *M* (*SD*), where the number outside the parentheses represents the mean and the number inside the parentheses represents the standard deviation.
The third column reports the means and standard deviations of the item-level CVR values for the necessity of the situation within each condition. The mean CVR for a set of items is identical to the CVI, which serves as an indicator of the overall content validity of the scale.

indicators remained generally strong across the temperature settings, with most values comparable to or higher than those of the MG and MG-ca groups.

Across prompt strategies, substantial improvement emerged with iterative prompt refinement. Prompt v0 performed worse than the MG-ca group, but Prompt v1 performed better than the MG-ca group and nearly reached the acceptable value of CVI, and Prompt v2 surpassed the 0.70 threshold and produced higher scores on the other three indicators, indicating overall superior content validity.

### 4.5.3. *Comparison of content validity across different temperature settings*

The differences in content validity among Temp0.7, Temp0.9, Temp1.0, Temp1.1, MG and MG-ca were examined. According to the results of the Kruskal-Wallis H test (presented in Table S3 and Table S4 in the Supplementary Materials), significant differences were found among different temperature conditions and MG in terms of the rationality of options, $\chi^2(4) = 17.15$, $p = .002$, the rationality of scoring, $\chi^2(4) = 10.35$, $p = .035$, and the overall evaluation of item quality, $\chi^2(4) = 10.89$, $p = .028$, but not in the necessity of the situation, $\chi^2(4) = 8.29$, $p = .082$. Dunn's post hoc tests with Bonferroni correction were used to follow up these findings. The results indicated that the items in MG scored significantly lower than those under Temp0.7 and Temp1.0 in terms of the rationality of options, but no significant pairwise differences were found among different item groups on the other two indicators, as shown in Fig. 2. No significant difference was found among different temperature conditions and MG-ca in terms of the necessity of the situation, $\chi^2(4) = 2.42$, $p = .658$, the rationality of options, $\chi^2(4) = 6.71$, $p = .152$, the rationality of scoring, $\chi^2(4) = 2.18$, $p = .703$, and the overall evaluation of item quality, $\chi^2(4) = 1.69$, $p = .792$. The distribution of the data is presented in Fig. 3. Temp1.0 received higher mean ranks in terms of the necessity of the situation and the rationality of options, indicating that this temperature setting generated context descriptions that were more relevant and necessary to the target traits. However, at a temperature of 1.1, notable option scoring errors occurred, where three options received a score of 1 point at one of the items.

### 4.5.4. *Comparison of content validity across different prompt strategies*

The differences in content validity among Prompt v0, Prompt v1, Prompt v2, MG and MG-ca were examined. According to the results of the Kruskal-Wallis H test (presented in Table S5 and Table S6 in the Supplementary Materials), significant differences were found among the groups in terms of the necessity of the situation, $\chi^2(3) = 9.85$, $p = .020$, the rationality of options, $\chi^2(3) = 12.36$, $p = .006$, the rationality of scoring, $\chi^2(3) = 10.24$, $p = .017$, and the overall evaluation of item quality, $\chi^2(3) = 15.40$, $p = .002$, when comparing different prompt conditions and MG. Significant differences were also found among different temperature conditions and MG-ca in terms of the necessity of the situation, $\chi^2(3) = 8.76$, $p = .033$, but not in the rationality of options, $\chi^2(3) = 6.02$, $p = .111$, the rationality of scoring, $\chi^2(3) = 4.05$, $p = .256$, and the overall evaluation of item quality, $\chi^2(3) = 6.91$, $p = .075$. Dunn's post hoc tests with Bonferroni correction were used to follow up these findings, revealing statistically significant differences in the rationality of options between Prompt v1/v2 and MG and overall item quality between Prompt v1 and MG, as shown in Fig. 4. Significant differences were also found between Prompt v0 and Prompt v2 in terms of the necessity of the situation when comparing different prompt conditions to MG-ca, as shown in Fig. 5. Both Prompt v1 and Prompt v2 consistently achieved higher mean ranks across all four indicators compared to Prompt v0, MG and MG-ca. Notably, option scoring errors were also observed in the Prompt v0 condition, where three options receiving a score of 0 points at one of the items. This error was mainly due to the fact that only example items were provided in Prompt v0 without incorporating explicit scoring format specifications.

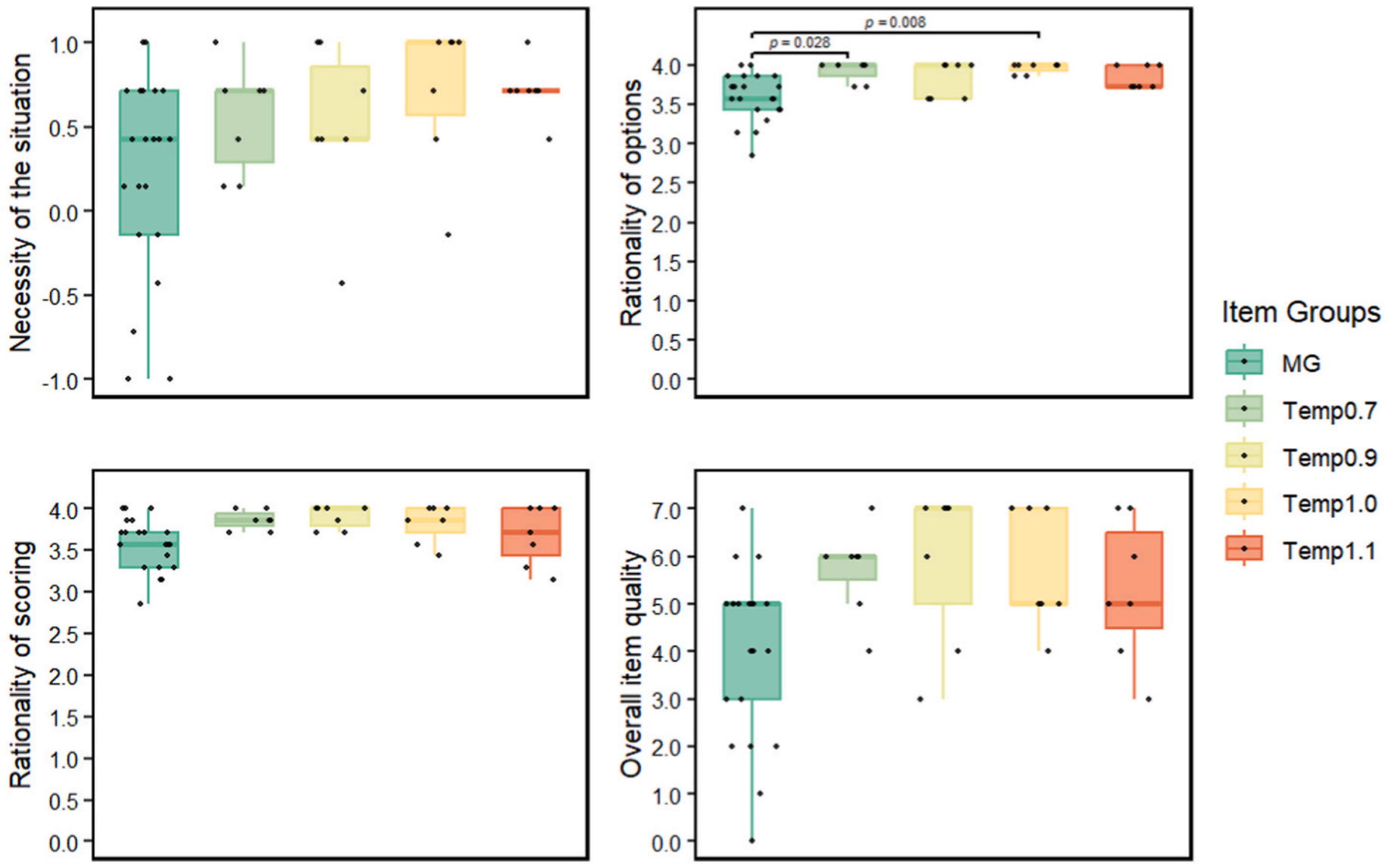

**Fig. 2.** Boxplots and scatter plots of content validity indicators of different temperature settings and MG.

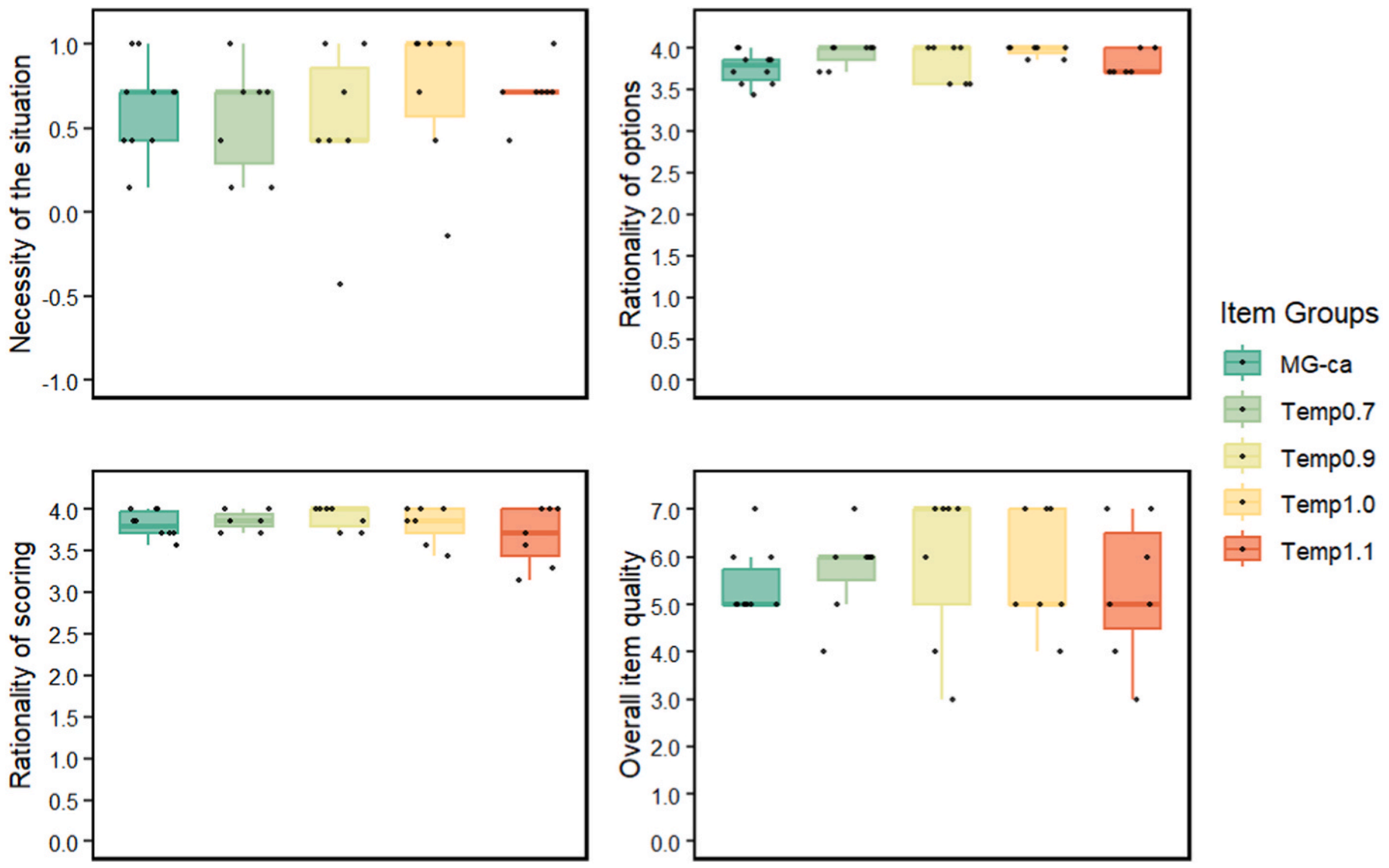

**Fig. 3.** Boxplots and scatter plots of content validity indicators of different temperature settings and MG-ca.

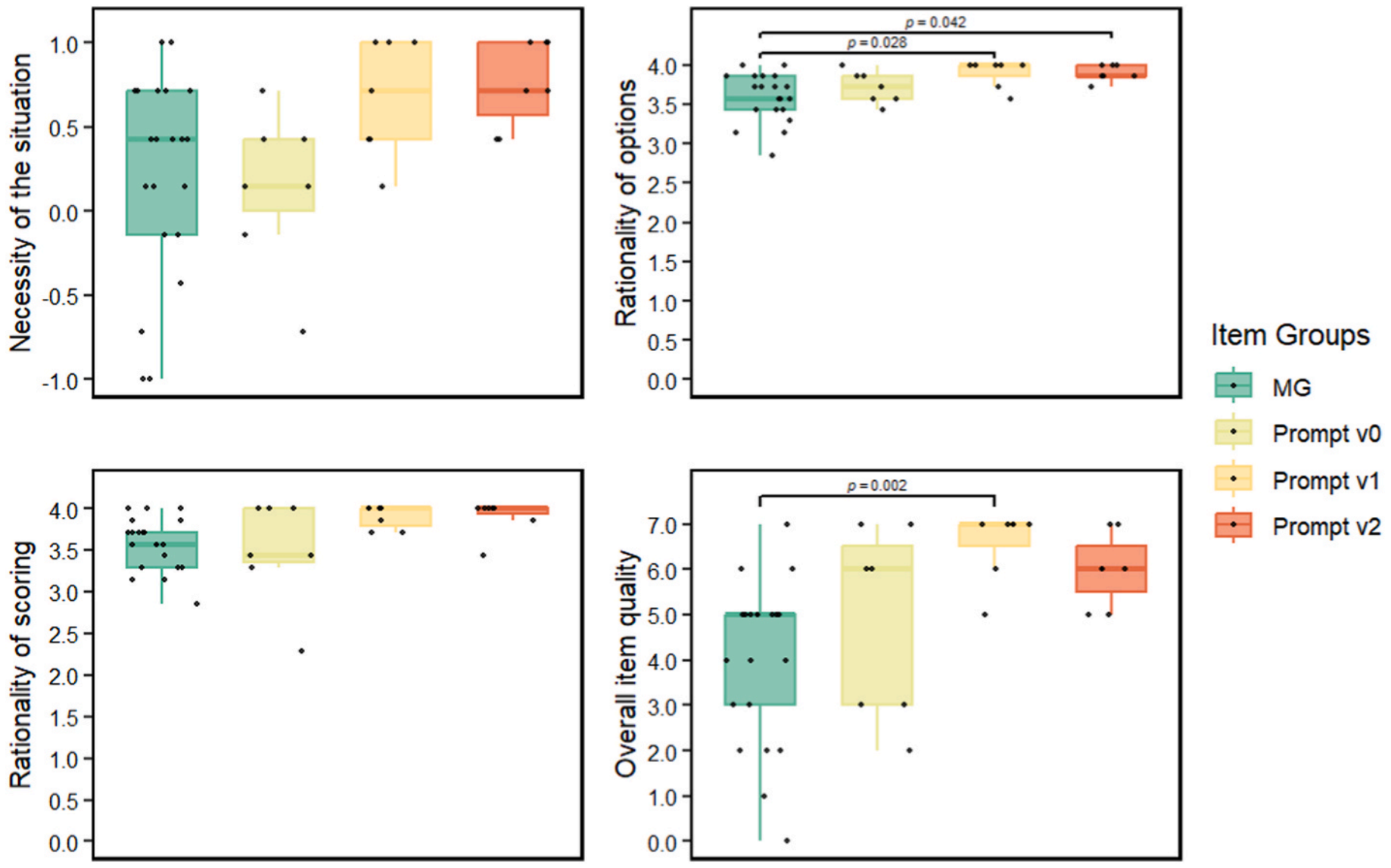

**Fig. 4.** Boxplots and scatter plots of content validity indicators of different prompt strategies and MG.

*4.5.5. Comparison of content validity at different times*

The results of the Mann-Whitney U test, presented in Table S7 in the Supplementary Materials and Fig. 6, indicated no significant differences between the items of Time1 and Time2 in terms of the necessity of the situation, $U = 21.50$, $z = -0.41$, $p = .710$, the rationality of options, $U = 22.50$, $z = -0.32$, $p = .805$, the rationality of scoring, $U = 29.00$, $z = 0.62$, $p = .620$, and the overall evaluation of item quality, $U = 34.00$, $z = 1.36$, $p = .259$. These findings suggest that the content validity of the items generated using the prompt v1 and a temperature of 1.0 exhibits temporal stability.

*4.6. Discussion*

Study 1 aimed to develop and evaluate an effective and reliable LLM-based AIG approach to SJT generation. Focusing on comparing different prompt strategies and temperature parameters, the study addressed three research questions that collectively demonstrated the effectiveness and stability of the proposed approach. First, items generated at a temperature setting of 1.0 achieved the highest content validity, striking an optimal balance between creativity and text quality. Second, items generated using optimized prompts (Prompt v1 and v2) significantly outperformed those generated with unoptimized prompts (Prompt v0). Although improvements made in Prompt v2 resulted in minor, non-significant gains, it is noteworthy that the CVI of Prompt v2 reached 0.76, the highest among all conditions. In contrast, the prompt without an explicit scoring format (i.e. Prompt v0) and high temperature setting (i.e. temperature = 1.1) introduced inaccuracies in option scoring. Third, no significant differences in content validity were found across items generated at different times, indicating the temporal stability of LLM-generated items.

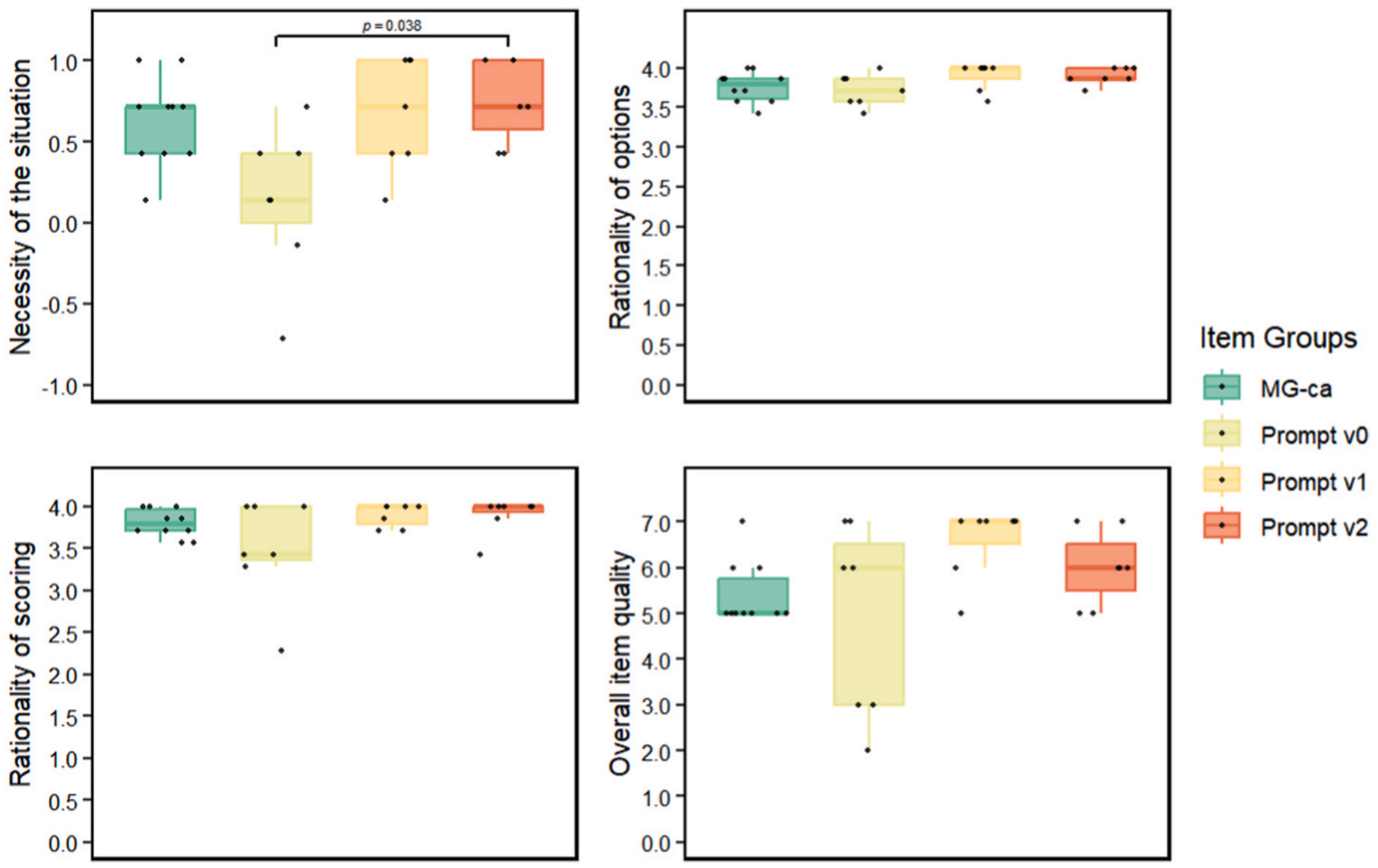

**Fig. 5.** Boxplots and scatter plots of content validity indicators of different prompt strategies and MG-ca.

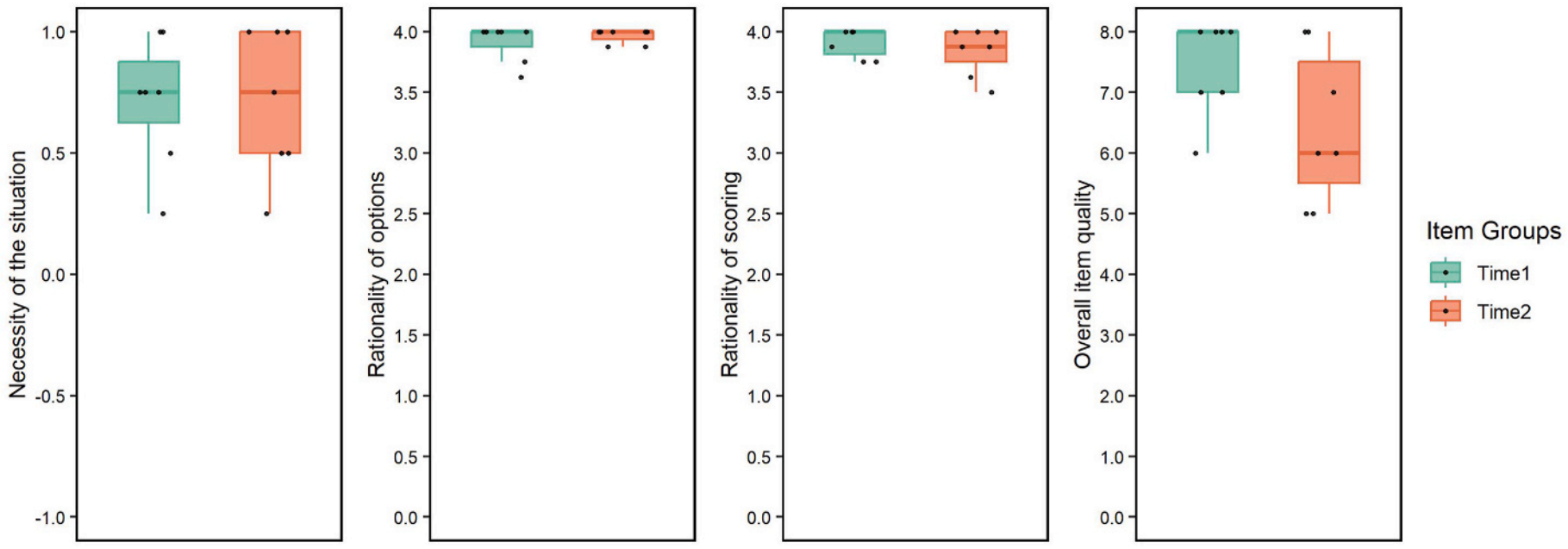

**Fig. 6.** Boxplots and scatter plots of content validity indicators at different times.

Study 1 demonstrates that, with optimized prompts and temperature settings (i.e., Prompt v2 combined with a temperature of 1.0), LLMs are capable of producing SJTs with content validity comparable to manually-generated SJT items. These findings demonstrate key methodological principles that support structured and reliable SJT generation using LLM-based AIG approaches.

## 5. Study 2: content validity across five personality domains and multiple rounds of generation

Study 2 aimed to further evaluate the robustness of the LLM-based AIG approach validated in Study 1 by examining its generalizability across personality traits and models and its reproducibility across multiple rounds of item generation. Building on the procedures established previously, this study implemented the method using the web-based ChatGPT-5 interface (2025, November 6 version, instant mode, https://chatgpt.com/) and optimized prompts (Prompt v2), without altering any default system settings. In this study, we generated items targeting one facet within each of the Big Five personality domains. The five facets are self-consciousness (from neuroticism), gregariousness (from extraversion), openness to ideas (from openness to experience), compliance (from agreeableness), and self-discipline (from conscientiousness). There was no item editing or selection conducted in the item generation procedures. To evaluate the reproducibility of the proposed approach, Study 2 examined whether the content validity of 200 LLM-generated items, produced across five consecutive rounds, varied as a function of either the generation round or the targeted facet.

### 5.1. Participants

The expert panel consisted of seven doctoral students in psychology from Beijing Normal University, all of whom were familiar with the development of personality tests and situational judgment tests.

### 5.2. Materials

#### 5.2.1. Personality situational judgment test generated by ChatGPT-5

SJT items for the five personality facets were generated by ChatGPT-5. Following the item generation procedure developed in Study 1, we employed the validated optimal prompt framework (Prompt v2) and conducted five consecutive rounds of item generation. In each round, ChatGPT-5 produced items for all five facets, with eight items per facet. The prompts were entered sequentially within the same conversation: after completing the items for one facet, the prompt for the next facet was submitted.

To minimize potential order effects, the presentation sequence of the five facets was determined by a Williams Latin square design. This design guarantees that each facet appears in every ordinal position exactly once across the five rounds, and provides a near-balance in the pairwise presentation order of all facets. Importantly, while each round was completed within a single conversation, different rounds were generated in independent conversations to prevent cross-round contamination. In total, 200 items were produced across the five rounds.

### 5.3. Procedures

A panel of seven experts rated the content validity using the identical four indicators in Study 1. The experts were not informed that the items were generated by ChatGPT-5. Prior to evaluation, they were provided with the definitions of each personality facet as well as detailed explanations of the four indicators and their corresponding rating criteria. Then they were asked to evaluate each item based on four indicators.

During the evaluation process, items were not presented in the order of generation rounds. Instead, they were organized by facet, such that each expert first rated all 40 items for self-consciousness, followed by the items for gregariousness, openness to ideas, compliance, and self-discipline. Within each facet, items appeared in the original order in which they were generated, and the presentation order was fixed and identical across all experts.

### 5.4. Statistical analysis

To examine whether the proposed approach demonstrates cross-model generalizability, we calculated the CVI for all items generated by ChatGPT-5 separately for each of the five rounds and each of the five facets, and then compared these values with the CVIs obtained in Study 1 from items generated using GPT-4 with the optimized parameters (Prompt v2 and a temperature of 1.0).

To further determine whether the method can be reliably applied across different personality facets and whether it can consistently produce high-quality items across multiple rounds, we treated facet and generation round as independent variables and the four indicators of content validity as dependent variables. Using IBM SPSS Statistics (Version 26), we conducted a series of two-way independent-measures ANOVAs to assess the effects of facet and round on each content validity indicator. If significant differences were found, post hoc tests were performed with Bonferroni error correction.

Inter-rater reliability was assessed using the same ICC(2,k) calculation method as described in Study 1.

### 5.5. Results

#### 5.5.1. Inter-rater reliability of expert ratings

Inter-rater reliability analysis revealed moderate to good agreement in the necessity of the situation (0.77), the rationality of options (0.54), the rationality of scoring (0.64), and the overall evaluation of item quality (0.65), according to Koo and Li's (2016) suggested thresholds for moderate (0.50 $<$ ICC $<$0.75) and good (0.75 $<$ ICC $<$0.90) reliability.

#### 5.5.2. Descriptive statistics of content validity across models, facets, and rounds

Table 2 presents the descriptive statistics for the four content validity indicators across models, facets, and rounds. Using the critical value of CVI $\geq$ 0.70 proposed by Tilden et al. (1990) as the criterion for acceptable content validity, both the CVI for items generated by GPT-4 (0.76) and ChatGPT-5 (0.71) exceeded this threshold, suggesting that the proposed approach generalizes reasonably well across different GPT model versions.

Across facets, CVI values ranged from 0.60 to 0.84. Three facets—Openness to ideas, Self-discipline, and Self-consciousness—clearly exceeded the 0.70 threshold, indicating strong content validity. Gregariousness (0.66) fell slightly below the cutoff but remained close to it, whereas Compliance showed the lowest value (0.60), indicating comparatively weaker content validity.

Across generation rounds, CVI values demonstrated considerable stability: four of the five rounds produced CVIs between 0.69 and 0.80, with only Round 4 (0.63) falling notably below the 0.70 threshold. The absence of systematic drift across rounds suggests that ChatGPT-5 maintained consistent performance, and that the stochastic nature of LLMs did not substantially undermine the reliability of the generation process.

#### 5.5.3. Differences in content validity across facets and rounds

A series of two-way ANOVAs were conducted to examine the effects of personality facet, generation round, and their interaction on the four indicators of content validity. Prior to the main analysis, the assumptions of ANOVA were evaluated. Shapiro-Wilk tests indicated that the data for all four indicators did not meet the assumption of normality ($ps < 0.001$), exhibiting distinct negative skewness. Given that the classical mean-based Levene's test is sensitive to non-normality, we assessed the homogeneity of variance using the Brown-Forsythe test (based on the median), as recommended by Brown and Forsythe (1974) for robust evaluation against non-normal distributions. The results indicated that the assumption of homogeneity of variance was met for the factor of generation round across all four indicators ($ps > 0.05$). Regarding the factor of personality facet, the assumption was met for the rationality of options and the overall evaluation of item quality ($ps > 0.05$), whereas significant heterogeneity of variance was observed for the necessity of

**Table 2**
Means and standard deviations of content validity indicators across models, facets, and rounds.

| Condition | Number of items | Necessity of the situation | Rationality of options | Rationality of scoring | Overall item quality |
|---|---|---|---|---|---|
| Items generated by ChatGPT-5 across facets | | | | | |
| Self-consciousness (N) | 40 | 0.71 (0.34) | 3.87 (0.22) | 3.75 (0.33) | 5.78 (1.23) |
| Gregariousness (E) | 40 | 0.66 (0.44) | 3.88 (0.28) | 3.87 (0.27) | 5.90 (1.53) |
| Openness to ideas (O) | 40 | 0.84 (0.26) | 3.93 (0.16) | 3.88 (0.21) | 6.13 (1.07) |
| Compliance (A) | 40 | 0.60 (0.53) | 3.88 (0.22) | 3.66 (0.54) | 5.43 (2.00) |
| Self-discipline (C) | 40 | 0.73 (0.40) | 3.89 (0.20) | 3.73 (0.37) | 5.65 (1.51) |
| Items generated by ChatGPT-5 across rounds | | | | | |
| Round 1 | 40 | 0.65 (0.51) | 3.87 (0.27) | 3.73 (0.51) | 5.75 (1.85) |
| Round 2 | 40 | 0.77 (0.31) | 3.95 (0.09) | 3.86 (0.26) | 5.93 (1.23) |
| Round 3 | 40 | 0.69 (0.37) | 3.90 (0.17) | 3.78 (0.29) | 5.68 (1.44) |
| Round 4 | 40 | 0.63 (0.48) | 3.84 (0.32) | 3.69 (0.44) | 5.55 (1.75) |
| Round 5 | 40 | 0.80 (0.32) | 3.89 (0.17) | 3.84 (0.25) | 5.98 (1.14) |
| All items generated by ChatGPT-5 | 200 | 0.71 (0.41) | 3.89 (0.22) | 3.78 (0.37) | 5.77 (1.50) |
| Items generated by GPT-4 with Prompt v2 and temperature = 1.0 | 7 | 0.76 (0.26) | 3.90 (0.11) | 3.90 (0.21) | 6.00 (0.82) |

*Note.* Values in the last four columns are presented in the format *M* (*SD*), where the number outside the parentheses represents the mean and the number inside the parentheses represents the standard deviation.
The third column reports the means and standard deviations of the item-level CVR values for the necessity of the situation within each condition. The mean CVR for a set of items is identical to the CVI, which serves as an indicator of the overall content validity of the scale.

the situation and the rationality of scoring ($ps < 0.05$). Given that the assumption of homogeneity was satisfied in the majority of conditions and considering the robustness of ANOVA against violations of normality and their relative resilience to variance heterogeneity in balanced designs with equal sample sizes ($N = 8$ per cell in the current study; Glass et al., 1972), we proceeded with the standard parametric two-way ANOVA.

The results of the two-way ANOVAs showed that the main effect of facet was significant only for the rationality of scoring, $F(4,175) = 2.83$, $p = .026$, $\eta_p^2 = 0.06$, whereas no significant effects emerged for the other three indicators: the necessity of the situation, $F(4,175) = 1.97$, $p = .101$, $\eta_p^2 = 0.04$; the rationality of options, $F(4,175) = 0.39$, $p = .819$, $\eta_p^2 = 0.01$; and the overall evaluation of item quality, $F(4,175) = 1.24$, $p = .296$, $\eta_p^2 = 0.03$. Post hoc comparisons indicated that none of the pairwise differences of the rationality of scoring among the five facets reached statistical significance.

The main effect of round was not significant across all four indicators: the necessity of the situation, $F(4,175) = 1.36$, $p = .250$, $\eta_p^2 = 0.03$; the rationality of options, $F(4,175) = 1.41$, $p = .232$, $\eta_p^2 = 0.03$; the rationality of scoring, $F(4,175) = 1.57$, $p = .185$, $\eta_p^2 = 0.03$; and the overall evaluation of item quality, $F(4,175) = 0.55$, $p = .696$, $\eta_p^2 = 0.01$.

Similarly, the interaction between facet and round was not significant for any indicator: the necessity of the situation, $F(16,175) = 0.68$, $p = .809$, $\eta_p^2 = 0.06$; the rationality of options, $F(16,175) = 1.02$, $p = .435$, $\eta_p^2 = 0.09$; the rationality of scoring, $F(16,175) = 0.89$, $p = .585$, $\eta_p^2 = 0.07$; and the overall evaluation of item quality, $F(16,175) = 1.19$, $p = .279$, $\eta_p^2 = 0.10$.

To further ensure the robustness of our results, we also conducted the Scheirer-Ray-Hare test, a non-parametric alternative for two-way ANOVA (Scheirer et al., 1976). The non-parametric analyses yielded consistent results, with no significant main effects or interactions observed. This confirms that our conclusions are robust to the choice of statistical method. The detailed results of the tests for normality and homogeneity of variance, and the non-parametric analyses are available in the Supplementary Materials (Table S8–S10). Together, these results indicate that neither the facet being measured nor the generation round meaningfully influenced the content validity of the items.

### 5.6. Discussion

Study 2 was designed to extend the methodological evaluation conducted in Study 1 by examining whether the LLM-based AIG approach exhibits cross-model generalizability, applicability across personality domains, and reproducibility across multiple rounds of generation. Consistent with the findings of Study 1, Study 2 demonstrated that the proposed LLM-based AIG procedure generalized successfully to ChatGPT-5. Both GPT-4 (from Study 1) and ChatGPT-5 produced items with CVIs above the established threshold of 0.70, suggesting that the approach does not rely on properties of a specific model.

Study 2 examined whether the method performs consistently across facets drawn from the Big Five personality domains. Although the items measuring gregariousness fell slightly below the cutoff and those assessing compliance displayed comparatively weaker CVI values, inferential analyses indicated no significant differences across facets for three of the four validity indicators. Even for the rationality of scoring, where a small facet effect emerged, pairwise comparisons showed no significant differences between any two facets. These results suggest that the proposed LLM-based AIG approach can be applied across different personality traits with reasonable content validity.

Another central aim of Study 2 was to test whether repeated item generation would lead to fluctuations in item quality due to the inherent randomness of LLMs. The results provide clear evidence of methodological reproducibility. Across five independent rounds, four rounds produced CVIs above or very close to 0.70 threshold, and no significant differences across rounds emerged in any of the four validity indicators. These findings demonstrate that, when guided by optimized prompts and a structured procedure, LLMs can repeatedly generate items with stable content validity, even when multiple rounds are performed in separate conversations.

Taken together, the findings of Study 2 provide strong support for the robustness of the LLM-based AIG procedure developed in Study 1. The method demonstrated cross-model generalizability, cross-domain applicability, and repeatable item quality, all of which are essential for establishing LLM-based AIG as a credible tool in psychological assessment.

## 6. Study 3: reliability and validity of LLM-generated personality situational judgment tests

Study 3 was conducted to demonstrate the practical application and utility of the proposed methodology by examining whether LLM-generated SJTs (LLM-SJTs) can serve as a fully functional personality assessment. Using GPT-4 with the optimized parameters identified in Study 1 (Prompt v2 and temperature of 1.0), we developed a LLM-SJT covering all five dimensions of the Big Five personality traits. In contrast to Study 1 and Study 2, which focused on expert-based content validation, Study 3 evaluated the practical psychometric performance of the LLM-SJT through empirical data collected from the target population.

### 6.1. Participants

A total of 468 participants were recruited across three stages of data collection. Across all stages, inclusion criteria for valid responses were: (1) age between 18 and 60 years; (2) all attention check items answered correctly; and (3) an average response time above 2 s per item. In the first stage, 24 participants were excluded for failing the attention checks and one for being under the age of 18. In the second stages, the Credemo platform's automatic verification function ensured that all responses met the inclusion criteria. In total, 443 valid participants completed the LLM-SJT and NEO-PI-R, 130 participants completed the criterion questionnaire, and 80 participants completed the LLM-SJT retest.

After applying the inclusion criteria, 443 valid questionnaires (264 females, 59.6%) were retained. Among them, 97.5% held a bachelor's degree or higher, and their ages ranged from 18 to 58 years, with a mean age of 28.25 years ($SD = 7.92$). All 443 participants completed the LLM-SJT and the NEO-PI-R facet items. A subsample of 130 college students (67 females, 51.5%) additionally completed the criterion measures for criterion validation. Their ages ranged from 18 to 28, with a mean age of 21.12 years ($SD = 1.92$). Another subsample of 80 participants (51 females, 63.8%) completed the two-week retest. Their ages ranged from 20 to 53, with a mean age of 31.60 years ($SD = 6.40$).

### 6.2. Materials

#### 6.2.1. Personality situational judgment test generated by GPT-4

Using OpenAI's "gpt-4-1106-preview" model, with optimized prompts (Prompt v2) and a temperature of 1.0, a LLM-SJT was generated to measure five facets of the Big Five personality traits. The five facets are identical to Study 2. For each facet, eight items were generated, resulting in a total of 40 items. Each item featured four response options, with two options reflecting high levels of the trait scoring 1 point each and the other two representing low levels of the trait scoring 0 points. The score for each facet was determined by calculating the sum of the items. There was no item editing or selection conducted in the item generation procedures.

#### 6.2.2. NEO-PI-R scale

The same five facets were selected from the NEO-PI-R scale (Costa & McCrae, 2008). Each facet consists of eight items, resulting in a total of 40 items. A 5-point Likert scale was utilized for scoring (1 for *strongly* *disagree* and 5 for *strongly agree*). The Cronbach's α coefficients of the five facets of NEO-PI-R in this study are 0.72 for self-consciousness, 0.88 for gregariousness, 0.85 for openness to ideas, 0.58 for compliance, and 0.89 for self-discipline.

#### 6.2.3. *Criterion measures*

Five constructs were selected as criteria, all of which have been shown to have significant correlations with one or more dimensions of the Big Five personality traits (e.g., Jones et al., 2011; Paulhus & Williams, 2002; Strauman, 2002). The criterion measures included five items measuring subjective well-being (Diener et al., 1985), five items measuring depression (Norton, 2007), seven items measuring game addiction (Lemmens et al., 2009), four items measuring aggression (Buss & Perry, 1992), and 12 items measuring the Dark Triad personality traits (Jonason & Webster, 2010). A total of 33 items were included, all employing a 5-point Likert scale for scoring (1 for *strongly disagree* and 5 for *strongly agree*). The Cronbach's α coefficients of the criterion measures in this study are 0.84 for subjective well-being, 0.89 for depression, 0.91 for game addiction, 0.63 for aggression, 0.76 for Machiavellianism of Dark Triad personality, 0.74 for psychopathy of Dark Triad personality, and 0.85 for narcissism of Dark Triad personality.

### 6.3. *Procedures*

Participants were recruited and questionnaires were distributed through social media and Credemo, a questionnaire and crowdsourcing platform. Each participant received approximately 10 Chinese Yuan (around 1.4 US dollars) as compensation.

Data collection was conducted in three stages. In the first stage, 338 participants completed three demographic questions, the LLM-SJT and NEO-PI-R items personality facets (40 items each), and three attention check items designed to detect careless responding. In the second stage, for the purpose of criterion-related validation, 130 additional participants were recruited to complete the same LLM-SJT and NEO-PI-R items, along with a criterion questionnaire (33 items in total). In the third stage, to assess test-retest reliability, the same group of 338 participants in the first stage was contacted two weeks later, and 80 participants voluntarily completed the LLM-SJT retest. Participants recruited in the first stage did not complete the criterion measures due to constraints related to survey length and participant burden.

### 6.4. *Statistical analysis*

Item analysis, reliability analysis (internal consistency and test-retest reliability), and validity analysis (convergent, discriminant, and criterion-related validity) were conducted using IBM SPSS Statistics (Version 26). A confirmatory factor analysis (CFA) using Mplus (Version 8.3; Muthén & Muthén, 1998-2019) was conducted for both the LLM-SJT and the NEO-PI-R. Given that the item scores for both the LLM-SJT and the NEO-PI-R are ordered-categorical variables, the Robust Weighted Least Squares (WLSMV) estimation techniques were used, as recommended by Flora and Curran (2004).

### 6.5. *Results*

#### 6.5.1. *Item analysis*

The item-total correlations were calculated to evaluate the item discrimination under each facet. The LLM-SJT items for self-consciousness (ranging from 0.21 to 0.57), gregariousness (ranging from 0.39 to 0.67), and openness to ideas (ranging from 0.32 to 0.50) exhibited relatively high discrimination. In contrast, the items for compliance (ranging from 0.17 to 0.33) and self-discipline (ranging from 0.22 to 0.38) showed lower discrimination.

#### 6.5.2. *Reliability*

The Cronbach's α coefficients were calculated as an indicator of internal consistency reliability (presented in Table 3). The Cronbach's α coefficients of the subscales of LLM-SJT measuring self-consciousness, gregariousness, and openness to ideas demonstrated acceptable internal consistency. However, the coefficients for compliance and self-discipline fell below the average values of 0.68, which was obtained from a meta-analysis of 271 SJTs by Kasten and Freund (2015). The LLM-SJT contained eight items per facet, whereas the manually-generated SJT by Mussel et al. (2018) included 22 items per facet. Despite having fewer items, the Cronbach's α coefficients for all five facets in the LLM-SJT were equal to or higher than those reported by Mussel et al. (2018).

The Pearson correlations were calculated to assess the test-retest reliability. Before conducting these analyses, we compared participants who completed only the pretest ($N = 233$) with those who completed both the pretest and retest sessions ($N = 80$) on demographic variables and five facet scores of both LLM-SJT and NEO-PI-R. No significant differences were found, indicating that participants who returned for the retest did not systematically differ from those who did not. Based on this, we proceeded to estimate test-retest reliability and found acceptable test-retest reliability for the five facets (presented in Table 3). However, the Pearson correlations for the facet of openness to ideas indicated relatively lower test-retest reliability.

#### 6.5.3. *Confirmatory factor analysis*

The five-factor structure of the LLM-SJT and the NEO-PI-R was examined using CFA. Model fit indices are presented in Table 4. Except for SRMR, all other model fit indicators of the LLM-SJT were better than those of the NEO-PI-R.

#### 6.5.4. *Convergent and discriminant validity*

Convergent validity was assessed by calculating the Pearson correlations between the corresponding facets of the LLM-SJT and NEO-PI-R. Discriminant validity was evaluated by computing the absolute values of the Pearson correlations between different facets of the LLM-SJT and the NEO-PI-R. The results are shown in Table 5. Correlations between corresponding facets of the LLM-SJT and NEO-PI-R ranged from 0.36 to 0.68, with a mean of 0.53 ($SD = 0.14$), showing solid evidence for convergent validity across most facets. However, the facet of compliance showed poor convergent validity. The correlation between LLM-SJT and NEO-PI-R in the facet of compliance was only 0.36, which was even lower than the correlations between the compliance facet of LLM-SJT and gregariousness and openness to ideas facets of NEO-PI-R. The average absolute value of the correlations between different facets in the LLM-SJT was 0.38, and that in the NEO-PI-R was 0.50, indicating that discriminant validity for the LLM-SJT was superior to that of the NEO-PI-R.

#### 6.5.5. *Criterion-related validity*

The Pearson correlations between the facets of LLM-SJT and NEO-PI-R and criterion measures were calculated. The results are shown in Table 6.

The LLM-SJT and NEO-PI-R demonstrated consistent directions in the correlations between personality dimensions and external criteria. In the self-consciousness facet, both LLM-SJT and NEO-PI-R showed

**Table 3**
Internal consistency and test-retest reliability of five facets of the LLM-SJT.

| LLM-SJT facet | Internal consistency reliability | Test-retest reliability |
|---|---|---|
| Self-consciousness (N) | 0.75 | 0.58 |
| Gregariousness (E) | 0.84 | 0.77 |
| Openness to ideas (O) | 0.70 | 0.41 |
| Compliance (A) | 0.57 | 0.52 |
| Self-discipline (C) | 0.61 | 0.65 |

**Table 4**
The model fit indices for the LLM-SJT and NEO-PI-R.

| Measurement | $\chi^2$ | *df* | *p* | $\chi^2/df$ | RMSEA | CFI | TLI | SRMR |
|---|---|---|---|---|---|---|---|---|
| LLM-SJT | 968.85 | 730 | <0.001 | 1.33 | 0.03 | 0.95 | 0.95 | 0.11 |
| NEO-PI-R | 1968.58 | 730 | <0.001 | 2.70 | 0.06 | 0.94 | 0.93 | 0.06 |

**Table 5**
The correlation matrix of the facets of LLM-SJT and NEO-PI-R.

| | Facet | 1 | 2 | 3 | 4 | 5 | 6 | 7 | 8 | 9 |
|---|---|---|---|---|---|---|---|---|---|---|
| 1 | LLM-SJT self-consciousness | | | | | | | | | |
| 2 | LLM-SJT gregariousness | *−.48* | | | | | | | | |
| 3 | LLM-SJT openness to ideas | *−.48* | *.39* | | | | | | | |
| 4 | LLM-SJT compliance | *−.30* | *.33* | *.44* | | | | | | |
| 5 | LLM-SJT self-discipline | *−.40* | *.32* | *.40* | *.30* | | | | | |
| 6 | NEO-PI-R self-consciousness | **.68** | −0.59 | −0.40 | −0.33 | −0.38 | | | | |
| 7 | NEO-PI-R gregariousness | −0.53 | **.67** | 0.46 | 0.39 | 0.31 | *−.66* | | | |
| 8 | NEO-PI-R openness to ideas | −0.46 | 0.57 | **.48** | 0.39 | 0.33 | *−.59* | *.60* | | |
| 9 | NEO-PI-R compliance | −0.30 | 0.32 | 0.16 | **.36** | 0.21 | *−.38* | *.36* | *.29* | |
| 10 | NEO-PI-R self-discipline | −0.52 | 0.49 | 0.33 | 0.34 | **.44** | *−.65* | *.54* | *.47* | *.42* |

*Note.* Numbers in bold indicate the evidence for convergent validity. Numbers in italic indicate the evidence for discriminant validity.
All the *p*-values of correlation coefficients in this table are smaller than 0.001.

**Table 6**
The Pearson correlations between the facets of LLM-SJT and NEO-PI-R and criterion measures.

| Facet | SWB | DE | GA | AG | DT-MA | DT-PA | DT-NA |
|---|---|---|---|---|---|---|---|
| LLM-SJT self-consciousness | −0.07 | 0.25 | 0.15 | 0.29 | 0.17 | 0.07 | 0.11 |
| *p* | 0.456 | 0.003 | 0.082 | 0.001 | 0.059 | 0.466 | 0.235 |
| LLM-SJT gregariousness | 0.04 | −0.23 | −0.15 | −0.11 | −0.11 | −0.06 | 0.11 |
| *p* | 0.645 | 0.010 | 0.081 | 0.227 | 0.220 | 0.466 | 0.224 |
| LLM-SJT openness to ideas | −0.11 | −0.13 | −0.12 | −0.08 | −0.08 | −0.16 | 0.17 |
| *p* | 0.202 | 0.146 | 0.163 | 0.396 | 0.382 | 0.077 | 0.054 |
| LLM-SJT compliance | 0.06 | −0.28 | −0.17 | −0.16 | −0.14 | −0.31 | 0.05 |
| *p* | 0.533 | 0.001 | 0.057 | 0.076 | 0.112 | <0.001 | 0.551 |
| LLM-SJT self-discipline | −0.04 | −0.16 | −0.28 | −0.22 | −0.24 | −0.12 | −0.15 |
| *p* | 0.655 | 0.066 | 0.001 | 0.011 | 0.007 | 0.167 | 0.099 |
| NEO-PI-R self-consciousness | −0.35 | 0.41 | 0.34 | 0.32 | 0.23 | 0.06 | 0.22 |
| *p* | <0.001 | <0.001 | <0.001 | <0.001 | 0.008 | 0.501 | 0.012 |
| NEO-PI-R gregariousness | −0.03 | −0.36 | −0.10 | −0.08 | −0.06 | −0.16 | 0.17 |
| *p* | 0.733 | <0.001 | 0.241 | 0.378 | 0.485 | 0.070 | 0.047 |
| NEO-PI-R openness to ideas | 0.08 | −0.15 | 0.01 | 0.11 | −0.02 | 0.01 | 0.03 |
| *p* | 0.398 | 0.085 | 0.958 | 0.231 | 0.783 | 0.940 | 0.758 |
| NEO-PI-R compliance | 0.13 | −0.17 | −0.21 | −0.50 | −0.27 | −0.34 | −0.25 |
| *p* | 0.137 | 0.052 | 0.019 | <0.001 | 0.002 | <0.001 | 0.005 |
| NEO-PI-R self-discipline | 0.37 | −0.38 | −0.47 | −0.35 | −0.33 | −0.25 | −0.11 |
| *p* | <0.001 | <0.001 | <0.001 | <0.001 | <0.001 | 0.004 | 0.202 |

*Note.* SWB = subjective well-being, DE = depression, GA = game addiction, AG = aggression, DT-MA = Machiavellianism of Dark Triad personality, DT-PA = psychopathy of Dark Triad personality, DT-NA = narcissism of Dark Triad personality.

significant positive correlations with depression and aggression. In the gregariousness facet, both tests revealed significant negative correlations with depression. In the openness to ideas facet, neither LLM-SJT nor NEO-PI-R exhibited significant correlations with the external criteria. In the compliance facet, both tests showed significant negative correlations with psychopathy. In the self-discipline facet, both tests demonstrated negative correlations with game addiction, aggression, and Machiavellianism. Although the NEO-PI-R showed higher criterion-related validity, this may be attributable to the shared method of measurement (i.e., Likert-type questionnaires) used for both the NEO-PI-R and the criterion measures.

### 6.6. Discussion

Previous studies have shown that situational judgment tests often face the psychometric problem of poor construct validity (Weekley et al., 2006; Whetzel & McDaniel, 2009). However, in this study, we demonstrated that the LLM-SJT exhibited satisfactory reliability and validity across multiple psychometric evaluations, offering psychometric properties comparable to traditional, manually developed Likert scales. These findings provide empirical support for the final step of our automated SJT generation approach, showing that the proposed methodological framework is not only theoretically sound but also practically capable of producing valid and reliable SJTs suitable for applied research and assessment contexts.

The observation that the relatively low convergent validity between the LLM-SJT and NEO-PI-R on the compliance facet and the lower criterion-related validity of the LLM-SJT compared to the NEO-PI-R does not necessarily indicate deficiencies in the LLM-SJT itself. Rather, it may partly reflect methodological constraints arising from the use of Likert-type scales as the sole validation criterion. Because SJTs and Likert-type scales differ both in the way items are presented and in the response processes they elicit, they may capture different aspects of the same construct.

From the perspective of measurement format, while Likert-type scales typically measure self-reported trait levels through abstract statements, SJTs assess behavioral choices in context-specific scenarios, which may more closely approximate real-life decision-making and situational tendencies (Kwon & Lee, 2020). From the perspective of construct measurement, prior research has suggested that SJTs may in fact measure general domain knowledge (Lievens & Motowidlo, 2016) or common sense (Salter & Highhouse, 2009). According to these theories, SJT performance reflects individuals' knowledge about the effectiveness of trait-related behaviors in specific situations. People who score higher on a trait not only tend to exhibit corresponding behaviors but also possess greater knowledge of when these behaviors are useful, leading them to endorse such responses in SJTs. Thus, the relatively small to moderate correlations often observed between SJTs and personality questionnaires (e.g., Christian et al., 2010) may be explained by the fact that SJTs capture implicit knowledge about trait utility rather than trait levels per se.

Taken together, these considerations suggest that the moderate convergent validity and criterion-related validity of the LLM-SJT may be less a sign of poor measurement quality and more an indication that SJTs and Likert-type scales tap into partially distinct psychological processes. Future research should disentangle this issue by incorporating behavioral outcomes or other real-world indicators as validation criteria, thereby providing a stronger test of the LLM-SJT's capacity to predict trait-relevant behavior beyond the constraints of Likert-type scales.

## 7. General discussion

This study systematically investigated four core questions: (1) how temperature settings influence the content validity of LLM-generated SJT items, (2) how prompt strategies can be optimized to enhance the content validity of LLM-generated SJT items, (3) whether the proposed approach produces reproducible items across multiple rounds and different model versions, and (4) whether LLM-generated items produced under the proposed approach can form a fully functional scale that demonstrates satisfactory reliability and validity and performs comparably to traditional, well-established measurement instruments. By optimizing temperature settings and prompt strategies, a structured and generalizable methodology for automated SJT item generation was developed and validated. Subsequent analyses thoroughly evaluated both cross-sectional psychometric evidence and the stability of generated items across multiple rounds and different models, thereby offering a more comprehensive validation framework for the proposed approach.

As part of our commitment to open science, we have made the full item bank and the full text of prompts generated and used in this research, a brief guide to using GPT-4 API, and anonymized raw data, along with analysis scripts, publicly available via the OSF repository (https://osf.io/jbvq7/). This will enable other researchers to replicate our findings, adapt the materials to different cultural contexts, and extend this line of research.

### *7.1. Feasibility and efficiency of LLM-based AIG in SJT generation*

Traditional situational judgment test development is time-consuming and labor-intensive and relies heavily on experts' personal experience, which can introduce bias and inconsistency (Hernandez & Nie, 2023). The application of LLMs to streamline test development holds significant potential for improving both efficiency and quality. This study demonstrated notable advancements in test generation efficiency and quality compared to previous research.

Regarding efficiency, prior studies such as Hernandez and Nie (2023) utilized GPT-2 to generate a large pool of personality items. Nevertheless, the process they implemented was laborious and time-consuming, requiring the filtering of over a million generated items. Similarly, Götz et al. (2024) relied on GPT-2 to generate 1000 items, from which only 20 met the desired psychometric standards, yielding a 2% success rate. In contrast, the test items generated in the current study have demonstrated solid psychometric properties without requiring extensive manual filtering. This improvement not only reduces the overall time required for test development but also cuts costs, making effective test generation accessible even in resource-limited settings. Compared to traditional SJT development, which often requires weeks or even months of expert involvement, the proposed generation process produced a complete item set within just a few minutes. On average, the cost of developing a written SJT for a specific job within a specific organization was estimated at $60,000 to $120,000 in 2008 (Lievens et al., 2008), and given inflation and rising personnel costs, the current expense is likely to be even higher. By contrast, based on OpenAI's API token pricing, generating 100 SJT items with GPT models costs approximately 0.02 to 1.00 US dollars, depending on the specific model used. This dramatic reduction in both time and financial cost underscores the practical innovation of our approach and highlights the potential of LLMs to substantially streamline SJT development.

### *7.2. Cultural adaptation and expert review in LLM-generated SJTs*

In the current study, LLMs successfully generated a personality situational judgment test in Chinese, filling the gap left by previous studies that focused predominantly on English-language tests. Notably, in Study 1, the manually-generated SJT used as a comparison was translated from German into Chinese. However, while cultural adaptation procedures were implemented during translation, the absence of formal cross-cultural scale validation might have retained certain linguistic and cultural discrepancies. These residual differences could partially account for the relatively lower content validity observed in the manually translated version compared to the LLM-generated items, as evidenced by the enhanced content validity following systematic removal of culturally incongruent manually-generated items. The observed cultural adaptation of LLM-generated items may stem from its training corpus containing more contemporary sociolinguistic patterns compared to translated items rooted in source-culture schemas, aligning with findings on LLMs' emergent cultural sensitivity (Pawar et al., 2024; Zhao, Huang, & Li, 2024). The findings tentatively suggest that LLM-generated items may exhibit better cultural appropriateness than conventional translation approaches under these specific conditions. Future research should further explore the generalizability of this approach across diverse cultural contexts, including formal validation processes to better disentangle the effects of cultural adaptation and AI generation.

While the integration of LLMs in SJT generation process demonstrates substantial advantages in operational efficiency and cross-cultural adaptation potential, these technological advantages do not obviate the critical need for subject matter experts (SMEs) in ensuring content validity. As evidenced in Study 1, even with the improved prompts (Prompt v1 and v2) and a temperature of 1.0, seven out of 21 LLM-generated items were rated by SMEs as having suboptimal overall quality (ratings <6 out of 7), and 7 item was rated as failing to reflect the target personality traits (CVR < 0.71). A similar pattern was observed in Study 2, where 29.5% of the 200 items received an overall quality rating below 6, and 25% of the items obtained a CVR score below 0.71 threshold. Though this proportion remained substantially lower than that observed for the translated manually-generated items, where 19 of the 21 items received overall quality ratings below 6 and 14 obtained CVR scores below 0.71 threshold, it underscores the necessity for systematic expert review. As recommended by Oeljeklaus et al. (2025), items generated through this methodology should be proceeded with the standard scale development procedure, requiring systematic evaluation, revision, and selection by SMEs to ensure comprehensibility, readability, and alignment with intended measurement objectives.

### *7.3. Optimizing prompt strategies for enhanced test quality*

Prompting has emerged as a crucial interface in human-computer interaction (Qiao et al., 2023), with some experts even highlighting it as a critical skill for the 21st century (Warzel, 2023). While previous studies (e.g., Hernandez & Nie, 2023; Hommel et al., 2022) explored the potential of fine-tuning language models for automatic item generation, the process requires extensive training data, which is often scarce for niche test types like personality situational judgment tests. In response to this limitation, P. Lee et al. (2023) employed prompt-based techniques instead of fine-tuning with promising psychometric results.

The present study optimized prompt strategies and developed a framework specifically tailored for generating SJTs. Results from Study 1 further clarify which prompt strategies were most effective in enhancing test quality. Specifically, prompts that incorporated precise construct definitions, detailed context and output specifications, Chain-of-Thought reasoning, expert persona adoption, and emotional prompting helped LLMs generate richer, context-specific scenarios and avoid scoring errors observed in less designed prompts. These findings underscore that well-designed prompts, drawing on multiple strategies in combination, are critical for aligning LLM-generated items with the psychometric standards required for SJTs.

Future research can leverage this prompt framework to tailor situational judgment tests to different personality dimensions, task objectives, or contextual constraints. Moreover, given the general applicability of prompts across various LLMs, researchers may implement this framework with other LLMs, thus enhancing its utility across platforms. If the framework requires adaptation for different tasks, it can be modified using the prompt strategies outlined in this study.

### *7.4. Limitations and future research*

To enrich the scenarios in the LLM-SJT, this study designed the prompts to generate content that included both everyday life and workplace situations, which participants found engaging and interesting. However, some participants in the study were undergraduate students, many of whom may lack real-world work experience. This lack of experience could affect their ability to fully understand and assess work-related scenarios, potentially impacting the validity and generalizability of the test results. Therefore, the current results are limited to student or general samples, and future research should consider incorporating scenarios that are more aligned with the life experience of the target population or recruiting actual applicant samples to enhance ecological validity.

Another potential avenue for future research is extending the LLM-SJT to cover all Big Five personality dimensions comprehensively. This study focused on five facets of the Big Five personality, demonstrating the feasibility and validity of using LLMs to generate SJTs. Building on this, future research could aim to develop and validate situational judgment tests that encompass all dimensions of the Big Five. By leveraging the generative capabilities of LLMs, researchers can create engaging and contextually diverse items for each personality dimension, leading to the development of comprehensive personality situational judgment tests.

Additionally, future studies could explore customizing situational judgment tests for specific populations or contexts, using LLMs to generate scenario-based assessments that are more relatable and engaging for particular groups. Integrating storytelling elements into SJTs may also enhance test validity, engagement, and interaction, providing a more immersive experience (Landers & Collmus, 2022).

## 8. Conclusion

This study developed and evaluated a structured and generalizable framework for automatically generating personality SJTs. Through systematic investigation of temperature settings, prompt strategies, and the reproducibility of generated items across models and generation rounds, we demonstrated that the proposed LLM-based AIG approach can produce conceptually coherent and psychometrically sound SJT items with substantially reduced time and resource demands.

Using the optimized methodological components, we further generated a Big Five personality SJT and showed that the resulting LLM-SJT demonstrated satisfactory reliability and validity. These findings indicate that a carefully designed LLM-based AIG framework can feasibly support the construction of fully functional assessment instruments that perform comparably to traditional, manually developed measurement instruments.

Overall, this research provides a methodological foundation for future work seeking to leverage LLMs in psychological assessment development. Continued investigation in applied settings and diverse populations will be essential for extending the practical utility and ecological validity of SJTs generated through LLM-based AIG approach.

## CRediT authorship contribution statement

**Chang-Jin Li:** Writing – review & editing, Visualization, Validation, Methodology, Formal analysis. **Jiyuan Zhang:** Writing – original draft, Project administration, Investigation, Formal analysis, Data curation. **Yun Tang:** Writing – review & editing, Investigation. **Jian Li:** Writing – review & editing, Supervision, Project administration, Methodology, Conceptualization.

## Ethics approval and consent

This study was approved by the Ethics Review Committee of the Faculty of Psychology at Beijing Normal University (IRB No. BNU202403130065) on March 13, 2024. All participants in this study provided informed consent via an online questionnaire. Prior to completing the questionnaire, participants were presented with the informed consent form and confirmed their consent by clicking the “Agree” button.

## Declaration of generative AI and AI-assisted technologies in the writing process

During the preparation of this work the authors used ChatGPT-5 in order to improve the readability and language of the manuscript. After using this tool, the authors reviewed and edited the content as needed and take full responsibility for the content of the publication.

## Funding

This research did not receive any specific grant from funding agencies in the public, commercial, or not-for-profit sectors.

## Declaration of competing interest

The authors declare that they have no known competing financial interests or personal relationships that could have appeared to influence the work reported in this paper.

## Appendix A. Supplementary data

Supplementary data to this article can be found online at https://doi.org/10.1016/j.chbr.2026.100964.

## Data availability

I have shared the link to my data at the Attach File step

Replication Materials for: Automatic Item Generation for Personality Situational Judgment Tests with Large Language Models (Original data) (OSF)

## References

API reference. OpenAI developer platform. (n.d.). https://platform.openai.com/docs/api-reference/chat.

Arendasy, M. (2005). Automatic generation of Rasch-calibrated items: Figural matrices test GEOM and endless-loops test EC. *International Journal of Testing, 5*(3), 197–224. https://doi.org/10.1207/s15327574ijt0503_2
Arendasy, M. E., Sommer, M., & Gittler, G. (2010). Combining automatic item generation and experimental designs to investigate the contribution of cognitive components to the gender difference in mental rotation. *Intelligence, 38*(5), 506–512. https://doi.org/10.1016/j.intell.2010.06.006
Attali, Y., Runge, A., LaFlair, G. T., Yancey, K., Goodwin, S., Park, Y., & Von Davier, A. A. (2022). The interactive reading task: Transformer-based automatic item generation. *Frontiers in Artificial Intelligence, 5*, Article 903077. https://doi.org/10.3389/frai.2022.903077
Baghestani, A. R., Ahmadi, F., Tanha, A., & Meshkat, M. (2019). Bayesian critical values for Lawshe's content validity ratio. *Measurement and Evaluation in Counseling and Development, 52*(1), 69–73. https://doi.org/10.1080/07481756.2017.1308227
Bejar, I. I., Lawless, R. R., Morley, M. E., Wagner, M. E., Bennett, R. E., & Revuelta, J. (2002). A feasibility study of on-the-fly item generation in adaptive testing. *ETS Research Report Series, 2002*(2). https://doi.org/10.1002/j.2333-8504.2002.tb01890.x. i-44.
Brown, M. B., & Forsythe, A. B. (1974). Robust tests for the equality of variances. *Journal of the American Statistical Association, 69*(346), 364–367. https://doi.org/10.1080/01621459.1974.10482955
Buss, A. H., & Perry, M. (1992). The aggression questionnaire. *Journal of Personality and Social Psychology, 63*(3), 452–459. https://doi.org/10.1037/0022-3514.63.3.452
Christian, M. S., Edwards, B. D., & Bradley, J. C. (2010). Situational judgment tests: Constructs assessed and a meta-analysis of their criterion-related validities. *Personnel Psychology, 63*(1), 83–117. https://doi.org/10.1111/j.1744-6570.2009.01163.x
Corstjens, J., Lievens, F., & Krumm, S. (2017). Situational judgment tests for selection. In H. Goldstein,, E. Pulakos,, J. Passmore,, & C. Semedo, (Eds.), *The wiley blackwell handbook of the psychology of recruitment, selection and employee retention* (pp. 226–246). Wiley-Blackwell.
Costa, P. T., Jr., & McCrae, R. R. (2008). The revised NEO personality inventory (NEO-PI-R). In G. J. Boyle, G. Matthews, & D. H. Saklofske (Eds.), *The SAGE handbook of personality theory and assessment: Personality measurement and testing* (Vol. 2, pp. 179–198). SAGE Publications.
Demszky, D., Yang, D., Yeager, D. S., Bryan, C. J., Clapper, M., Chandhok, S., Eichstaedt, J. C., Hecht, C., Jamieson, J., Johnson, M., Jones, M., Krettek-Cobb, D., Lai, L., JonesMitchell, N., Ong, D. C., Dweck, C. S., Gross, J. J., & Pennebaker, J. W. (2023). Using large language models in psychology. *Nature Reviews Psychology, 2*, 688–701. https://doi.org/10.1038/s44159-023-00241-5
Diener, E., Emmons, R. A., Larsen, R. J., & Griffin, S. (1985). The satisfaction with life scale. *Journal of Personality Assessment, 49*(1), 71–75. https://doi.org/10.1207/s15327752jpa4901_13
Doughty, J., Wan, Z., Bompelli, A., Qayum, J., Wang, T., Zhang, J., Zheng, Y., Doyle, A., Sridhar, P., Agarwal, A., Bogart, C., Keylor, E., Kultur, C., Šavelka, J., & Sakr, M. (2024). *A comparative study of AI-generated (GPT-4) and human-crafted MCQs in programming education*. In N. Herbert, & C. Seton (Eds.) (pp. 114–123). https://doi.org/10.1145/3636243.3636256
Embretson, S. E. (2005). Measuring human intelligence with artificial intelligence: Adaptive item generation. In R. J. Sternberg,, & J. E. Pretz, (Eds.), *Cognition and intelligence: Identifying the mechanisms of the mind* (pp. 251–267). Cambridge University Press.
Embretson, S. E., & Yang, X. (2007). Automatic item generation and cognitive psychology. In C. R. Rao, & S. Sinharay (Eds.), *Handbook of statistics: Psychometrics* (Vol. 26, pp. 747–768). Elsevier North Holland.
Flora, D. B., & Curran, P. J. (2004). An empirical evaluation of alternative methods of estimation for confirmatory factor analysis with ordinal data. *Psychological Methods, 9*(4), 466–491. https://doi.org/10.1037/1082-989x.9.4.466
Gierl, M. J., & Lai, H. (2012). The role of item models in automatic item generation. *International Journal of Testing, 12*(3), 273–298. https://doi.org/10.1080/15305058.2011.635830
Gierl, M. J., & Lai, H. (2015). Using automated processes to generate test items and their associated solutions and rationales to support formative feedback. *Interaction Design and Architecture(s), 25*, 9–20. https://doi.org/10.55612/s-5002-025-001
Glass, G. V., Peckham, P. D., & Sanders, J. R. (1972). Consequences of failure to meet assumptions underlying the fixed effects analyses of variance and covariance. *Review of Educational Research, 42*(3), 237–288. https://doi.org/10.3102/00346543042003237
Götz, F. M., Maertens, R., Loomba, S., & van der Linden, S. (2024). Let the algorithm speak: How to use neural networks for automatic item generation in psychological scale development. *Psychological Methods, 29*(3), 494–518. https://doi.org/10.1037/met0000540
Hernandez, I., & Nie, W. (2023). The AI-IP: Minimizing the guesswork of personality scale item development through artificial intelligence. *Personnel Psychology, 76*(4), 1011–1035. https://doi.org/10.1111/peps.12543
Hommel, B. E., Wollang, F.-J. M., Kotova, V., Zacher, H., & Schmukle, S. C. (2022). Transformer-based deep neural language modeling for construct-specific automatic item generation. *Psychometrika, 87*(2), 749–772. https://doi.org/10.1007/s11336-021-09823-9
Hooper, A. C., Cullen, M. J., & Sackett, P. R. (2006). Operational threats to the use of SJTs: Faking, coaching, and retesting issues. In J. A. Weekley,, & R. E. Ployhart, (Eds.), *Situational judgment tests: Theory, measurement, and application* (pp. 205–232). Lawrence Erlbaum Associates Publishers.
Huang, Y. A. N., & He, L. (2016). Automatic generation of short answer questions for reading comprehension assessment. *Natural Language Engineering, 22*(3), 457–489. https://doi.org/10.1017/S1351324915000455
Jonason, P. K., & Webster, G. D. (2010). The dirty dozen: A concise measure of the dark triad. *Psychological Assessment, 22*(2), 420–432. https://doi.org/10.1037/a0019265
Jones, S. E., Miller, J. D., & Lynam, D. R. (2011). Personality, antisocial behavior, and aggression: A meta-analytic review. *Journal of Criminal Justice, 39*(4), 329–337. https://doi.org/10.1016/J.JCRIMJUS.2011.03.004
Kasten, N., & Freund, P. A. (2015). A meta-analytical multilevel reliability generalization of situational judgment tests (SJTs). *European Journal of Psychological Assessment, 32*(3), 230–240. https://doi.org/10.1027/1015-5759/a000250
Kasten, N., Freund, P. A., & Staufenbiel, T. (2020). "sweet little lies": An in-depth analysis of faking behavior on situational judgment tests compared to personality questionnaires. *European Journal of Psychological Assessment, 36*(1), 136–148. https://doi.org/10.1027/1015-5759/a000479
Koo, T. K., & Li, M. Y. (2016). A guideline of selecting and reporting intraclass correlation coefficients for reliability research. *Journal of Chiropractic Medicine, 15*(2), 155–163. https://doi.org/10.1016/j.jcm.2016.02.012
Krumm, S., Thiel, A. M., Reznik, N., Freudenstein, J.-P., Schäpers, P., & Mussel, P. (2024). Creating a psychological test in a few seconds: Can ChatGPT develop a psychometrically sound situational judgment test? *European Journal of Psychological Assessment*. https://doi.org/10.1027/1015-5759/a000878. Advance online publication.
Kwon, E., & Lee, J. (2020). Validation of honesty test using situational judgment Test Format. *Korean Journal of Industrial and Organizational Psychology, 33*(4), 545–569. https://doi.org/10.24230/kjiop.v33i4.545-569
Landers, R. N., & Collmus, A. B. (2022). Gamifying a personality measure by converting it into a story: Convergence, incremental prediction, faking, and reactions. *International Journal of Selection and Assessment, 30*(1), 145–156. https://doi.org/10.1111/ijsa.12373
Lawshe, C. H. (1975). A quantitative approach to content validity. *Personnel Psychology, 28*(4), 563–575. https://doi.org/10.1111/j.1744-6570.1975.tb01393.x
Lee, P., Fyffe, S., Son, M., Jia, Z., & Yao, Z. (2023). A paradigm shift from "human writing" to "machine generation" in personality test development: An application of state-of-the-art natural language processing. *Journal of Business and Psychology, 38*(1), 163–190. https://doi.org/10.1007/s10869-022-09864-6
Lee, U., Jung, H., Jeon, Y., Sohn, Y., Hwang, W., Moon, J., & Kim, H. (2024). Few-shot is enough: Exploring ChatGPT prompt engineering method for automatic question generation in english education. *Education and Information Technologies, 29*(9), 11483–11515. https://doi.org/10.1007/s10639-023-12249-8
Lemmens, J. S., Valkenburg, P. M., & Peter, J. (2009). Development and validation of a game addiction scale for adolescents. *Media Psychology, 12*(1), 77–95. https://doi.org/10.1080/15213260802669458
Li, Z., & Zhang, H. (2008). Computerized automatic item generation: An overview. *Advances in Psychological Science, 16*(2), 348–352. https://journal.psych.ac.cn/xlkxjz/CN/Y2008/V16/I2/348.
Lievens, F., & Motowidlo, S. J. (2016). Situational judgment tests: From measures of situational judgment to measures of general domain knowledge. *Industrial and Organizational Psychology, 9*(1), 3–22. https://doi.org/10.1017/iop.2015.71
Lievens, F., Peeters, H., & Schollaert, E. (2008). Situational judgment tests: A review of recent research. *Personnel Review, 37*(4), 426–441. https://doi.org/10.1108/00483480810877598
Marvin, G., Hellen, N., Jjingo, D., & Nakatumba-Nabende, J. (2023). Prompt engineering in large language models. In Jacob, I.J., Piramuthu, S., & Falkowski-Gilski, P. (Eds.), *International conference on data intelligence and cognitive informatics* (pp. 387-402). Singapore: Springer Nature. 10.1007/978-981-99-7962-2_30.
Mussel, P., Gatzka, T., & Hewig, J. (2018). Situational judgment tests as an alternative measure for personality assessment. *European Journal of Psychological Assessment, 34*(5), 328–335. https://doi.org/10.1027/1015-5759/a000346
Muthén, L. K., & Muthén, B. O. (1998-2019). *Mplus User's Guide* (8th ed.). Muthén & Muthén.
Norton, P. J. (2007). Depression anxiety and stress scales (DASS-21): Psychometric analysis across four racial groups. *Anxiety, Stress & Coping, 20*(3), 253–265. https://doi.org/10.1080/10615800701309279
Oeljeklaus, L., Höft, S., & Danner, D. (2025). Comparing psychometric properties of expert-developed and AI-Generated personality scales: A proof-of-concept Study. *Psychological Test Adaptation and Development, 6*, 29–43. https://doi.org/10.1027/2698-1866/a000095
Olaru, G., Burrus, J., MacCann, C., Zaromb, F. M., Wilhelm, O., & Roberts, R. D. (2019). Situational judgment tests as a method for measuring personality: Development and validity evidence for a test of dependability. *PLoS One, 14*(2), Article e0211884. https://doi.org/10.1371/journal.pone.0211884
OpenAI, Achiam, J., Adler, S., Agarwal, S., Ahmad, L., Akkaya, I., Aleman, F. L., Almeida, D., Altenschmidt, J., Altman, S., Anadkat, S., Avila, R., Babuschkin, I., Balaji, S., Balcom, V., Baltescu, P., Bao, H., Bavarian, M., Belgum, J., … Zoph, B. (2024). *GPT-4 technical report. arXiv*. https://doi.org/10.48550/arXiv.2303.08774
Paulhus, D. L., & Williams, K. M. (2002). The Dark Triad of personality: Narcissism, machiavellianism, and psychopathy. *Journal of Research in Personality, 36*(6), 556–563. https://doi.org/10.1016/s0092-6566(02)00505-6
Qi, X. (2013). Guanxi, social capital theory and beyond: Toward a globalized social science. *British Journal of Sociology, 64*(2), 308–324. https://doi.org/10.1111/1468-4446.12019
Qiao, S., Ou, Y., Zhang, N., Chen, X., Yao, Y., Deng, S., Tan, C., Huang, F., & Chen, H. (2023). *Reasoning with language model prompting: A survey. arXiv*. https://doi.org/10.48550/arXiv.2212.09597
Rush, B. R., Rankin, D. C., & White, B. J. (2016). The impact of item-writing flaws and item complexity on examination item difficulty and discrimination value. *BMC Medical Education, 16*(1), 250. https://doi.org/10.1186/s12909-016-0773-3

Salter, N. P., & Highhouse, S. (2009). Assessing managers' common sense using situational judgment tests. *Management Decision, 47*(3), 392–398. https://doi.org/10.1108/00251740910946660

Scheirer, C. J., Ray, W. S., & Hare, N. (1976). The analysis of ranked data derived from completely randomized factorial designs. *Biometrics, 32*(2), 429–434. https://doi.org/10.2307/2529511

Schwartz, S. H. (2014). National culture as value orientations: Consequences of value differences and cultural distance. In V. Ginsburgh, & D. Throsby (Eds.), *Handbook of the economics of art and culture* (Vol. 2, pp. 547–586). Elsevier. https://doi.org/10.1016/b978-0-444-53776-8.00020-9.

Slaughter, J. E., Christian, M. S., Podsakoff, N. P., Sinar, E. F., & Lievens, F. (2014). On the limitations of using situational judgment tests to measure interpersonal skills: The moderating influence of employee anger. *Personnel Psychology, 67*(4), 847–885. https://doi.org/10.1111/peps.12056

Strauman, T. J. (2002). Self-regulation and depression. *Self and Identity, 1*(2), 151–157. https://doi.org/10.1080/152988602317319339

Tilden, V. P., Nelson, C. A., & May, B. A. (1990). Use of qualitative methods to enhance content validity. *Nursing Research, 39*(3), 172–175. https://doi.org/10.1097/00006199-199005000-00015

von Davier, M. (2018). Automated item generation with recurrent neural networks. *Psychometrika, 83*(4), 847–857. https://doi.org/10.1007/s11336-018-9608-y

Wang, L., Song, R., Guo, W., & Yang, H. (2024). Exploring prompt pattern for generative artificial intelligence in automatic question generation. *Interactive Learning Environments*, 1–26. https://doi.org/10.1080/10494820.2024.2412082

Warzel, C. (2023, February 8). The most important job skill of this century. *The Atlantic*. https://www.theatlantic.com/technology/archive/2023/02/openai-text-models-google-search-engine-bard-chatbot-chatgpt-prompt-writing/672991/.

Weekley, J. A., Ployhart, R. E., & Holtz, B. C. (2006). On the development of situational judgment tests: Issues in item development, scaling, and scoring. In J. A. Weekley, & R. E. Ployhart, (Eds.), *Situational judgment tests: Theory, measurement, and application* (pp. 157–182). Lawrence Erlbaum Associates Publishers.

Wei, X., & Li, Q. (2013). The Confucian value of harmony and its influence on Chinese social interaction. *Cross-Cultural Communication, 9*(1), 60–66. https://doi.org/10.3968/j.ccc.1923670020130901.12018

Wei, X., Zhang, K., Fu, X., & Wang, F. (2023). Honor culture and face culture: A comparison through the lens of the dignity, honor, and face cultural framework and Indigenous social theory. *Advances in Psychological Science, 31*(8), 1541–1552. https://doi.org/10.3724/sp.j.1042.2023.01541

Whetzel, D. L., & McDaniel, M. A. (2009). Situational judgment tests: An overview of current research. *Human Resource Management Review, 19*(3), 188–202. https://doi.org/10.1016/j.hrmr.2009.03.007

Zhang, Z., Tu, Z., Chen, Y., Xiao, X., Feng, Y., & Zhang, W. (2026). Automated item generation for personality assessment: Development and validation of large-language-model-derived HEXACO situational judgment tests. *Journal of Research in Personality, 120*, Article 104680. https://doi.org/10.1016/j.jrp.2025.104680